\documentclass[letterpaper,journal]{IEEEtran}
\usepackage{amsmath,amsfonts}
\usepackage{algorithm}
\usepackage{array}
\usepackage{makecell}
\ifCLASSOPTIONcompsoc
\usepackage[caption=false,font=normalsize,labelfont=sf,textfont=sf,subrefformat=parens,labelformat=parens]{subfig}
\else
\usepackage[caption=false,font=footnotesize,subrefformat=parens,labelformat=parens]{subfig}
\fi
\usepackage{textcomp}
\usepackage{changes}
\usepackage{stfloats}
\usepackage{url}
\usepackage{soul}
\usepackage{verbatim}
\usepackage{graphicx}
\usepackage{newtxtext}
\usepackage{cite}
\usepackage{booktabs}
\usepackage{multirow}
\usepackage{tabularx}
\usepackage{array}
\usepackage{multirow}
\usepackage{threeparttable}
\usepackage{pifont}

\usepackage{algorithm}
\usepackage{algpseudocode}
\algrenewcommand\algorithmicrequire{\textbf{Input:}}
\algrenewcommand\algorithmicensure{\textbf{Output:}}

\begin{document}

\title{Quantum-Assisted Memory-Efficient Training for Parameter-Intensive Wi-Fi-Based Human Activity Recognition}
\author{To Truong An,~\IEEEmembership{Student Member, IEEE}, Jie Zhang,~\IEEEmembership{Senior Member, IEEE}, Guolin Yin,~\IEEEmembership{Member, IEEE}, 
Junqing~Zhang,~\IEEEmembership{Senior~Member,~IEEE}, Yanjiao Li,~\IEEEmembership{Member, IEEE},
Trung Q. Duong,~\IEEEmembership{Fellow, IEEE}, and 
Simon~L.~Cotton,~\IEEEmembership{Fellow, IEEE}
        % <-this % stops a space
\thanks{This work was supported in part by the Cyber AI Hub Doctoral Training Program at the Centre for Secure Information Technologies (CSIT), Queen’s University Belfast, in part by the UK Government as part of the New Deal for Northern Ireland, administered through Innovate UK/UKRI, in part by the Canada Excellence Research Chair (CERC) Program CERC-2022-00109, in part by the Natural Sciences and Engineering Research Council of Canada (NSERC) CREATE program (grant number 596205-2025),  and in part by the UK Engineering and Physical Sciences Research Council (EPSRC) through the EPSRC Hub on All Spectrum Connectivity under Grant EP/X040569/1 and Grant EP/Y037197/1.}

\thanks{To Truong An, Jie Zhang, Guolin Yin, and Simon L. Cotton are with the Centre for Wireless Innovation, School of Electronics, Electrical Engineering and Computer Science, Queen’s University Belfast, Belfast BT7 1NN, UK (e-mail: ato01@qub.ac.uk, jie.zhang@qub.ac.uk, g.yin@qub.ac.uk, simon.cotton@qub.ac.uk).

Junqing~Zhang is with the School of Computer Science and Informatics, University of Liverpool, Liverpool, L69 3DR, U.K. (email: Junqing.Zhang@liverpool.ac.uk).

Yanjiao Li is with the Institute of Engineering Technology, University of Science and Technology Beijing, Beijing 100083, China. (e-mail: yanjiaoli@ustb.edu.cn).

Trung Q. Duong is with the Faculty of Engineering and Applied Science, Memorial University, St. John’s, NL A1C 5S7, Canada, and also with the Centre for Wireless Innovation, School of Electronics, Electrical Engineering and Computer Science, Queen’s University Belfast, Belfast BT7 1NN, UK (e-mail: tduong@mun.ca).
}
\thanks{Corresponding author is Trung Q. Duong.}
}

\maketitle
\begin{abstract}
Wi-Fi-based human activity recognition (HAR) has become an important part of integrated sensing and communications, paving the way for a range of context-aware services. However, most existing Wi-Fi-based HAR systems rely on deep learning (DL) models that are computationally and memory intensive in both training and inference, which poses significant challenges for real-world deployment. Conventional training requires simultaneous updates of millions of parameters, leading to prohibitive memory consumption. In this paper, we propose a novel quantum-assisted memory-efficient training framework (Q-MET) designed to improve efficiency in both training and inference. Q-MET utilizes a hybrid quantum classical neural network to indirectly generate parameters for HAR models, significantly reducing the trainable parameter count compared to direct optimization. To further support the deployment on resource-constrained devices, we integrate structured pruning during the training phase. Experimental results demonstrate that Q-MET achieves a 90\% to 95\% reduction in trainable parameters compared with conventional backpropagation-based DL training  while maintaining or even exceeding classical classification accuracy. Additionally, Q-MET supports lightweight inference through structured pruning, achieving 75\% to 85\% model sparsity with less than 2\% loss in classification accuracy. To the best of our knowledge, this work represents the first quantum-assisted approach to simultaneously tackle memory inefficiencies in both the training and inference stages of HAR systems.
\end{abstract}

\section{Introduction}
Over the last decade, the rapid proliferation of sensing technologies has revolutionized human activity recognition (HAR) across various fields, enabling applications ranging from smart home, elder care, and safety monitoring~\cite{11066259,10824863,10707238,10476327,10664573,9947336}.
Among these technologies, Wi-Fi sensing has emerged as a particularly cost-effective solution, leveraging existing infrastructure to extract rich environmental information without the need for dedicated hardware sensors. Recognizing this potential, the IEEE 802.11bf Task Group was formed to work on an amendment to the IEEE 802.11 standard to specifically deliver support for Wi-Fi sensing~\cite{10467185,11161858,10547188,10192291}. This is a major step toward enabling reliable, scalable, and deployable Wi-Fi sensing systems for HAR and related applications. 

Wi-Fi sensing typically relies on either received signal strength indicator (RSSI) or channel state information (CSI)\cite{8794643,9855454,9748867}. Although RSSI is widely accessible, it is coarse-grained and susceptible to environment fluctuations, limiting its stability and discriminative capability~\cite{10938090,8469082,zhang2022handgest}. In contrast, CSI captures fine-grained physical-layer properties such as amplitude and phase responses across subcarriers. This allows for robust characterization of subtle motions and environmental changes~\cite{9900419,7438932,11097060,8515041}. 

To exploit these high-dimensional features, deep learning (DL) models have become the standard for Wi-Fi-based HAR offering the ability to automatically learn discriminative representations from large-scale datasets~\cite{YANG2023100703,10756643,10681250}.
Numerous studies have investigated the robustness of DL-based HAR models, focusing on their classification accuracy, resilience to channel variations, and inference efficiency~\cite{11000292,10855560,10842509,11242148, 10666859}. However, the widespread adoption of DL in HAR faces a critical challenge, the significant memory requirements. While DL models are typically evaluated in environments with abundant resources, practical HAR deployment often targets resource-constrained edge platforms and commercial-off-the-shelf (COTS) devices.

In recent years, cloud computing has emerged as a promising approach for handling large-scale data processing and training DL models~\cite{9148862, 10038471,garcia2020cloud}. Due to the limited computational and energy resources of edge devices, the training of DL-based HAR models is often offloaded to cloud platforms to leverage scalable resources, while inference is typically performed either locally on devices or at the edge. However, cloud-centric approaches introduce latency, privacy concerns, and reliance on network connectivity, drawbacks that are often unacceptable for real-time applications~\cite{LENKA2025103915}. On-device learning and inference offer a solution to these issues but must operate within strict memory and power constraints~\cite{10980613}.

Although training can be offloaded to cloud infrastructures, reducing training-related memory costs remains a critical challenge. Training DL models requires significant memory for storing model parameters, optimizer states, intermediate activations, and gradients~\cite{3433727}. Specifically, the backpropagation mechanism underlying every parameter update involves the storage of layer-by-layer activations to compute gradients. Consequently, training requires a large memory footprint, substantially greater than the memory required during inference~\cite{5555}. 
As wireless networks increasingly host DL-based applications, from wireless sensing, physical-layer security, channel estimation, and related tasks~\cite{11016788, 11266879, 11197518,10024264,11192647}, this training memory overhead becomes a prohibitive barrier to practical deployment~\cite{10552143,9667414}. However, most existing studies prioritize inference optimization and leave the issue of inefficient training largely unaddressed.

For the purpose of reducing model complexity and deploying models in real-world applications, especially on resource-constrained devices, researchers have extensively studied model compression techniques, primarily targeting inference efficiency. One of the most widely used strategies is quantization, reducing precision in model parameters by converting high-precision representations, such as FP32 (32-bit floating point), to low-bit forms such as INT8 (8-bit integer) or INT4 (4-bit integer). This method significantly reduces both memory footprint and computational cost~\cite{8578384}. The other popular technique in model compression is pruning, which involves finding redundant or less-significant connections from the neural network~\cite{10330640}. The pruning algorithm keeps only the most important weights and pathways, which reduces computation with little or no impact on the performance.

Despite these advances, state-of-the-art HAR optimization focuses almost exclusively on the inference phase. These methods, such as quantization and pruning, typically require training full-scale, parameter heavy models before applying compression techniques, and optionally additional fine-tuning, which further increases the computational and memory costs during training~\cite{LENKA2025103915,11000292}. As a result, these approaches remain memory‑intensive and computationally expensive during the training stage, rendering them unsuitable for memory-limited platforms.

To bridge this gap, we explore an emerging paradigm of quantum-assisted machine learning. The QuantumTrain (QT) framework utilizes a hybrid quantum-classical neural network to generate parameters for classical DL models. Because the quantum circuit acts as a highly expressive generator using significantly fewer trainable parameters than the target classical network, QT offers a pathway to massive reductions in training memory. While QT has been applied to image classification~\cite{10821200}, large language models (LLMs)~\cite{liu2025a}, and physical layer authentication~\cite{truong2026rf_fingerprint}, it has not yet been adapted to the specific constraints of HAR, nor does it inherently address inference efficiency.

In this paper, we propose a novel Quantum-assisted Memory-Efficient Training (Q-MET) framework. The primary objective of Q-MET is to enhance efficiency in both the training and inference phases of HAR system by introducing a pruning-enhanced QT approach. Unlike traditional approaches, where pruning is applied after the model has been fully trained, and often requiring additional fine-tuning, introducing extra computational overhead. Our method applies structured pruning directly within the QT training process. This design allows Q-MET to eliminate redundant parameters as it learns, avoiding the inefficiencies of post-training pruning. 
Ultimately, Q-MET produces a pruned and compact DL model that is immediately suitable for on-device inference, making it especially valuable for deployment in real-time, resource-limited HAR applications. By addressing inefficiencies in both training and inference, Q-MET provides a comprehensive solution for scaling DL-based HAR to practical, low-resource environments. Our main contributions are highlighted as follows

\begin{itemize}
    \item We identify and address the open problem of training inefficiency in  HAR systems, which poses significant challenges in real-world deployments. To the best of our knowledge, this is the first work to explicitly target training efficiency in this domain.
    
    \item We extend the QT framework to HAR by designing a hybrid quantum classical neural network, consisting of a quantum neural network and a classical mapping network, to generate the parameters of DL-based sensing models efficiently.
    
    \item We propose Q-MET, a novel memory-efficient end-to-end training framework for HAR. In Q-MET, structured pruning is applied directly within the QT process, enabling the model to remove redundant parameters during training, resulting in compact models that are directly deployable for on-device inference without post-training compression.

    \item We evaluate the performance of the proposed approach by applying Q-MET to train a DL-based HAR model on two public datasets: Widar3.0~\cite{9516988} and UT-HAR~\cite{8067693}. We experimentally demonstrate that Q-MET enables the training of DL-based HAR models with up to 95\% fewer trainable parameters compared with standard backpropagation-based training, substantially reducing memory requirements and enabling efficient training in cost-sensitive cloud environments, while preserving or slightly enhancing model performance. Furthermore, Q-MET enables lightweight inference by producing highly sparse models through structured pruning, achieving 75\% to 85\% sparsity without significant performance degradation, which makes the resulting models well suited for resource-constrained HAR deployments.
\end{itemize} 
The remainder of the paper is organized as follows. Section II reviews related works on Wi-Fi-based HAR systems and the QT framework, and Section III introduces preliminaries on quantum computing and pruning. Section IV presents the on-device Wi-Fi-based HAR system, and Section V details the design of Q-MET. Section VI presents the experimental setup and highlights the key results, and Section VII concludes the paper. For convenience, the main symbols used throughout the paper are summarized in Table~\ref{notation}.
\begin{table}[t]
\centering
\caption{Summary of Notation}
\label{notation}
\begin{tabular}{|c|l|}
\hline
\textbf{Symbol} & \multicolumn{1}{c|}{\textbf{Definition}} \\ \hline
$U(\theta)$ & Parameterized unitary of the PQC \\ \hline
$U_L(\theta_L)$ & Parameterized unitary at the $L$-th layer \\ \hline
$|\psi(\theta)\rangle$ & Output state of the PQC \\ \hline
$p_\theta(z)$ & Probability of measurement outcome $z$ \\ \hline
$z$ & Possible measurement outcome \\ \hline
$H(n)$ & Channel frequency response (CFR) \\ \hline
$X(n)$ & Transmitted signal in the frequency domain \\ \hline
$W(n)$ & Additive white Gaussian noise (AWGN) \\ \hline
$|H_k|$ & Amplitude of the $k$-th subcarrier \\ \hline
$\angle H_k$ & Phase of the $k$-th subcarrier \\ \hline
$\theta_{PQC}$ & Trainable parameters of the PQC \\ \hline
$N_q$ & Number of qubits \\ \hline
$\Psi$ & Measurement probability vector of the PQC \\ \hline
$|\phi_k\rangle$ & $k$-th computational basis state \\ \hline
$\mathcal{P}$ & Number of basis states, $\mathcal{P}=2^{N_q}$ \\ \hline
$\mathcal{S}(\Psi)$ & Sinusoidal embedding of $\Psi$ \\ \hline
$S_0, S_1$ & Sinusoidal embedding vectors \\ \hline
$M_{\theta_m}$ & Mapping network with parameters $\theta_m$ \\ \hline
$\theta_m$ & Trainable parameters of the mapping network \\ \hline
$\Theta_m$ & Matrix of generated classical parameters \\ \hline
$n_G$ & Number of parameter groups \\ \hline
$n_B$ & Number of parameters per group \\ \hline
$\overrightarrow{\theta}_{\text{ResNet-18}}$ & Generated parameter vector of ResNet-18 \\ \hline
$\mathcal{Z}(\cdot)$ & Forward function of ResNet-18 \\ \hline
$\mathcal{L}_{CE}$ & Cross-entropy loss \\ \hline
$N_{\text{Batch}}$ & Batch size \\ \hline
$N_c$ & Number of output classes \\ \hline
$\eta$ & Learning rate \\ \hline
$C_{\text{ResNet-18}}$ & Total number of parameters in ResNet-18 \\ \hline
$C_{\text{Q-MET}}$ & Total trainable parameters of Q-MET \\ \hline
$\Delta\mathcal{C}$ & Parameter efficiency gain \\ \hline
$\text{score}(u;W)$ & LAMP importance score of weight $u$ \\ \hline
$p$ & Pruning ratio \\ \hline
$s$ & Sparsity level \\ \hline
$T_e$ & Average training time per epoch \\ \hline
\end{tabular}
\end{table}

\section{Related Work}
This section presents a review of prior research on Wi-Fi-based HAR systems, quantum machine learning and QT framework. Table~\ref{Comparison_of_related_studies} provides a comparative summary of the existing methods in relation to our proposed approach. As summarized in Table~\ref{Comparison_of_related_studies}, prior HAR compression methods~\cite{11000292,LENKA2025103915,9622251} target inference efficiency only, whereas the QuantumTrain family~\cite{10821200,11575585,liu2025a,10821046,10821103} targets training efficiency only. In contrast, Q-MET is, to the best of our knowledge, the first framework to jointly address both training- and inference-stage memory efficiency. Moreover, Q-MET is compatible with existing compression techniques, as the structured pruning used in this work can be replaced by alternative pruning or quantization methods.

\subsection{Wi-Fi-based Human Activity Recognition Systems}
Wi-Fi-based HAR systems are capable of obtaining contextual information about human activities by analyzing changes of wireless signals~\cite{abuhoureyah2025location,zhang2020integrated}. In recent years, DL-based methods have been widely used for Wi-Fi-based HAR. For instance, Li et al.~\cite{li2020wihf} first developed a domain-independent motion change pattern of gestures, and then proposed a novel dual-task DL framework for gesture recognition. Gu et al.~\cite{gu2022wigrunt} proposed WiGRUNT based on a dual-attention mechanism to enhance cross-domain gesture recognition. Zhang et al.~\cite{zhang2021csi} presented a modified graph-based few-shot learning approach, which leveraged complex-valued convolutions in the frequency domain followed by channel-wise attention for Wi-Fi-based HAR. To alleviate the complexity and computational overhead of conventional DL models while enhancing the robustness and generalization of wireless sensing systems, lightweight neural networks and online learning methods have been increasingly adopted in HAR. 
% Specifically, UN-2DCNN, a lightweight 2D-CNN with temporal skipping, channel attention, and uncertainty-aware feature scaling coupled with a dual-stage decision head, was proposed for HAR with memory and energy constraints of devices~\cite{miao2025lightweight}.
Additionally, modified online sequential learning algorithms were developed to enhance Wi-Fi-based contactless localization, enabling robust and efficient domain adaptation without retraining existing models~\cite{zhang2025leveraging,zhang2022online}. 

Several studies have investigated the use of quantization and pruning techniques to enhance model efficiency in HAR systems. In~\cite{LENKA2025103915}, the authors demonstrated that the post-training-quantization could reduce up to 72\% model size of DL-based HAR model, while achieving an accuracy degradation within 5\%. Alternatively, channel pruning was employed in the design of the lightweight Wi-Fi CSI human sensing system (LWiHS)~\cite{11000292}. 
This approach achieved a reduction of over 50\% in the number of parameters, with less than a 0.5\% reduction in accuracy. While in~\cite{9622251}, a hybrid pruning-quantization approach have achieved significant complexity reductions. First, post-training pruning was used to reduce the model’s complexity, followed by quantizing the already pruned weights into the INT8 format. This hybrid approach provided a 27\% reduction in both computational complexity and storage, with only a 2\% drop in classification accuracy. 
\subsection{Quantum Machine Learning and QuantumTrain Framework}
Quantum machine learning (QML) has demonstrated great potential in enhancing learning efficiency and handling complex, high-dimensional data~\cite{biamonte2017quantum,dunjko2016quantum,11278495}. QML takes advantage of key quantum properties, such as superposition and entanglement, to perform computations across many basis states in parallel. This parallelism opens up new possibilities for improving computational efficiency, especially when working with complex or high-dimensional data~\cite{lau2017quantum,henderson2020quanvolutional,10113742}. In QML model, data is commonly fed into quantum circuits through encoding schemes like gate-angle encoding, where data values control the rotation of quantum gates, or amplitude encoding, which maps information directly into the amplitudes of a quantum state~\cite{huang2021power}. Practical applications of QML have already emerged in several fields, including channel estimation~\cite{11168959}, healthcare monitoring~\cite{10974616}, natural language processing~\cite{PERALGARCIA2024124427}, and resource optimization~\cite{10806885}. However, QML is still in its early stages and faces several key challenges. One of the most pressing issues is the difficulty of encoding large datasets into quantum circuits, which is constrained by the limited number of available qubits and the shallow circuit depths allowed by the short coherence times of current quantum hardware~\cite{10889597}. In addition to these encoding limitations, many QML models require quantum hardware not only during training but also at the inference stage. This reliance can lead to inefficiencies, especially in time-sensitive applications such as real-time decision-making in autonomous systems.

To address these challenges, a novel technique called QT has been proposed, which is employed  to train classical neural network models. The core idea of the QT framework is to employ a quantum neural network (QNN) alongside a mapping mechanism to generate the weights of a classical neural network model.
This approach offers significant parameter reduction on a polylogarithmic scale, eliminates data encoding issues, and enables inference on purely classical computers~\cite{10821056}. The authors in~\cite{10821200} explored the QT framework for image classification. By using the QT framework to train convolutional neural networks (CNNs), the model achieves a test accuracy of 80.21\% on the MNIST dataset while utilizing only 10.8\% of the original parameter count. Similarly, the author in~\cite{11011133} also leveraged the QT framework to train the CNN for the task of deepfake audio detection.
The other studies~\cite{10821103,10821046, 10827567,liu2025a} have explored the potential of the QT framework in training various classical DL models, including reinforcement learning (RL), long short-term memory (LSTM), LLMs and federated learning (FL). These studies further demonstrate that the QT framework can be effectively applied across diverse classical DL architectures. 

To enhance the scalability of the QT framework, the authors in~\cite{10889597} improved the existing mapping network by replacing the conventional multi-layer perceptron (MLP) with a tensor network-based mapping model. While this approach offers more efficient parameter representation and lowers computational demands, it also introduces a trade-off in the form of significantly increased training time. Meanwhile, the study in~\cite{11575585} addresses the training time problem of the QT framework. The authors proposed a novel embedding function that employs sine and cosine functions restricted to a single complete cycle, enabling the encoding of quantum state information corresponding to the QNN’s output probabilities. By reducing the input dimensionality of the mapping network, this method lowers its computational load and consequently accelerates the training process of the QT framework.

\newcolumntype{L}[1]{>{\raggedright\arraybackslash}m{#1}}
\newcolumntype{C}[1]{>{\centering\arraybackslash}m{#1}}

% helper for wrapped multirow text
\newcommand{\AdvCell}[1]{%
  \begin{minipage}[c]{\linewidth}
  \raggedright
  #1
  \end{minipage}
}

\begin{table*}[!t]
\centering
\renewcommand{\arraystretch}{1.3}
\setlength{\tabcolsep}{4pt}

\begin{threeparttable}
\caption{Summary of related studies with our proposed study}
\label{Comparison_of_related_studies}

\begin{tabular}{|C{2.2cm}|C{1.5cm}|L{5.0cm}|L{8.0cm}|}
\hline
\textbf{Tasks} & \textbf{Studies} & \centering\arraybackslash\textbf{Key Idea} & \centering\arraybackslash\textbf{Advantages \& Limitations} \\
\hline

\multirow{4}{*}{\makecell[c]{HAR Accuracy\\Improvement}}
& \cite{li2020wihf}
& Dual-task DL with domain-independent motion features
& \multirow{4}{=}{\AdvCell{%
\ding{51} Achieves high recognition accuracy in both in-domain and cross-domain settings while improving processing efficiency\\
\ding{51} Supports adaptation to environmental changes without full retraining\\
\ding{55} Requires complex DL architectures with high computational overhead, making deployment on resource-constrained devices challenging
}} \\
\cline{2-3}
& \cite{gu2022wigrunt}
& ResNet-based spatiotemporal dual-attention learning with domain-independent gesture features
& \\
\cline{2-3}
& \cite{zhang2021csi}
& Graph-based few-shot learning with dual attention for CSI-based HAR
& \\
\cline{2-3}
& \cite{zhang2025leveraging,zhang2022online}
& Online and incremental learning for robust sensing in dynamic environments
& \\
\hline

\multirow{3}{*}{\makecell[c]{HAR Efficiency\\Improvement}}
& \cite{11000292}
& Channel pruning for efficient Wi-Fi CSI-based sensing
& \multirow{3}{=}{\AdvCell{%
\ding{51} Reduces model complexity and parameter count while maintaining good recognition performance, enabling deployment on resource-constrained devices\\
\ding{55} Primarily improves inference efficiency, with limited gains in training efficiency
}} \\
\cline{2-3}
& \cite{LENKA2025103915}
& Quantization for efficient on-device Wi-Fi sensing
& \\
\cline{2-3}
& \cite{9622251}
& Network pruning and quantization for efficient device-free wireless sensing
& \\
\hline

\multirow{4}{*}{\makecell[c]{QuantumTrain\\Framework}}
& \cite{10821200,11575585}
& QT-based training for CNNs
& \multirow{4}{=}{\AdvCell{%
\ding{51} Improves training efficiency while maintaining comparable accuracy with a significantly reduced number of trainable parameters\\
\ding{55} Focuses primarily on the training stage, with limited impact on inference efficiency
}} \\
\cline{2-3}
& \cite{liu2025a}
& QT-based training for LLMs
& \\
\cline{2-3}
& \cite{10821046}
& QT-based training for LSTMs
& \\
\cline{2-3}
& \cite{10821103}
& QT-based training for RL models
& \\
\hline

\makecell[c]{HAR Efficiency\\Improvement}
& Our Work
& Novel quantum-assisted memory-efficient training framework for HAR
& \AdvCell{%
\ding{51} Significantly improves training efficiency by reducing the number of trainable parameters and producing highly sparse models, enabling lightweight on-device inference without substantial performance loss
} \\
\hline

\end{tabular}

\begin{tablenotes}
\footnotesize
\item Note: \ding{51} indicates advantages and \ding{55} indicates limitations.
\end{tablenotes}

\end{threeparttable}
\end{table*}

\section{Preliminaries on Quantum Machine Learning and Pruning Techniques}
\subsection{Quantum Machine Learning}
QML refers to the application of quantum computing to machine learning tasks, utilizing quantum mechanical properties such as superposition and entanglement to achieve computational advantages~\cite{9076118}. By exploiting the high dimension Hilbert space, QML offers the potential to process complex data more efficiently than classical models~\cite{10757594, 11244103}. Among the various QML architectures, the parameterized quantum circuits (PQCs) have received significant attention.

A PQC consists of a sequence of quantum gates controlled by a set of tunable parameters $\theta = \left\{ \theta_{1}, \theta_{2}, \ldots, \theta_{n} \right\}$. These parameters are optimized to minimize a task-specific cost function, often using classical optimization algorithms in a hybrid quantum-classical loop. A PQC can be mathematically represented as
\begin{equation}
    U(\theta) = U_L(\theta_L) \ldots U_2(\theta_2) U_1(\theta_1),
\end{equation}
where $U_L(\theta_L)$ denotes the parameterized unitary operation applied at the $L$-th layer of the circuit.

After applying the parameterized unitary $U(\theta)$ to an initial quantum state, which is usually set as zero state $|0\rangle^{\otimes n}$, with $n$ is the number of qubits, the PQC outputs a quantum state $|\psi(\theta)\rangle = U(\theta)|0\rangle^{\otimes n}$. Upon measurement in the computational basis, this state collapses to a probability vector. This probability can be used to compute expectation values or feed into classical post-processing steps. The probability of obtaining the outcome $z$ is
\begin{equation}
    p_\theta(z) = |\langle z | \psi(\theta) \rangle|^2,
\end{equation}
where $z$ denotes the possible measurement outcomes. For a quantum system of $n$ qubits, there are $2^n$ possible outcomes. $| \psi(\theta) \rangle$ represents the output state of PQC before measurement. 
The inner product $\langle z \mid \psi(\theta) \rangle$ gives the complex probability amplitude of observing the outcome $z$ when measuring the state $|\psi(\theta)\rangle$. Finally, the squared magnitude $\left| \langle z \mid \psi(\theta) \rangle \right|^{2}$ yields the actual probability of obtaining the bit string $z$ upon measurement~\cite{11248844,Cerezo2021VQA}.

\subsection{Pruning}
In recent years, the size of DL models has grown exponentially~\cite{7780459, krizhevsky2012imagenet}, leading to increased computational cost and memory usage. These challenges hinder deployment on edge devices and embedded systems with limited on-chip resources, where memory, latency, and energy efficiency are critical constraints~\cite{10330640,10643325}. To mitigate this issue, pruning has been emerged as one of the most effective model compression techniques, reducing redundant parameters while preserving model performance.

Pruning techniques can be broadly categorized into unstructured pruning and structured pruning. Unstructured pruning removes specific weights from the network according to their size or contributions to the loss function. Preliminary works have found that unstructured pruning can considerably minimize the CNN parameters with minimal loss of accuracy~\cite{zhang2018systematic, 105555}. Nevertheless, due to the sparse and irregular weight matrices, they are not well-suited to hardware acceleration, and are ineffectual in inference~\cite{LIANG2021370}.
Structured pruning, on the other hand, prunes out entire architectural components (filters, channels, or layers), which forms more regular, dense matrices~\cite{9758156}. This regularity enables improved general-purpose hardware and DL accelerator support, making it more suitable for real-time inference on resource-constrained platforms~\cite{10330640,zhao2025hape}. Structured pruning is often further classified into filter pruning (removing whole convolutional filters) and channel pruning (removing feature map channels)~\cite{8237417,8237803}. While structured pruning may incur a slightly higher accuracy trade-off than unstructured methods, it offers superior hardware compatibility and practical computational speed-ups.

\section{On-Device Wi-Fi–Based Human Activity Recognition Systems}
Wi-Fi-based HAR utilizes the CSI, extracted from the communications link between a transmitter and receiver, to characterize human movements. The received Wi-Fi signal in the frequency domain can be described as
\begin{equation}
\label{received_signal_frequency_domain}
    Y(n) = H(n)\cdot X(n) + W(n),
\end{equation}
where $H(n), X(n)$ and $W(n)$  represent channel frequency response (CFR), transmitted signal, and additive white Gaussian noise (AWGN) in frequency domain, respectively.

In practical deployment, depending on the specific CSI extraction tool used, the receiver samples the signal in the time domain, and the CFR is subsequently estimated at the subcarrier level, capturing both amplitude attenuation and phase shifts in complex form~\cite{9667414,YANG2023100703}. The CFR of each subcarrier can be expressed as
\begin{equation}
    H_{k} = \left| H_{k} \right|e^{j\angle H_{k}}, 
\end{equation}
where $\left| H_{k} \right|$ and $\angle H_{k}$ denote the amplitude and phase of the $k$-th subcarrier, respectively.

\begin{figure}[!t]
\centering
\includegraphics[width=2.8in]{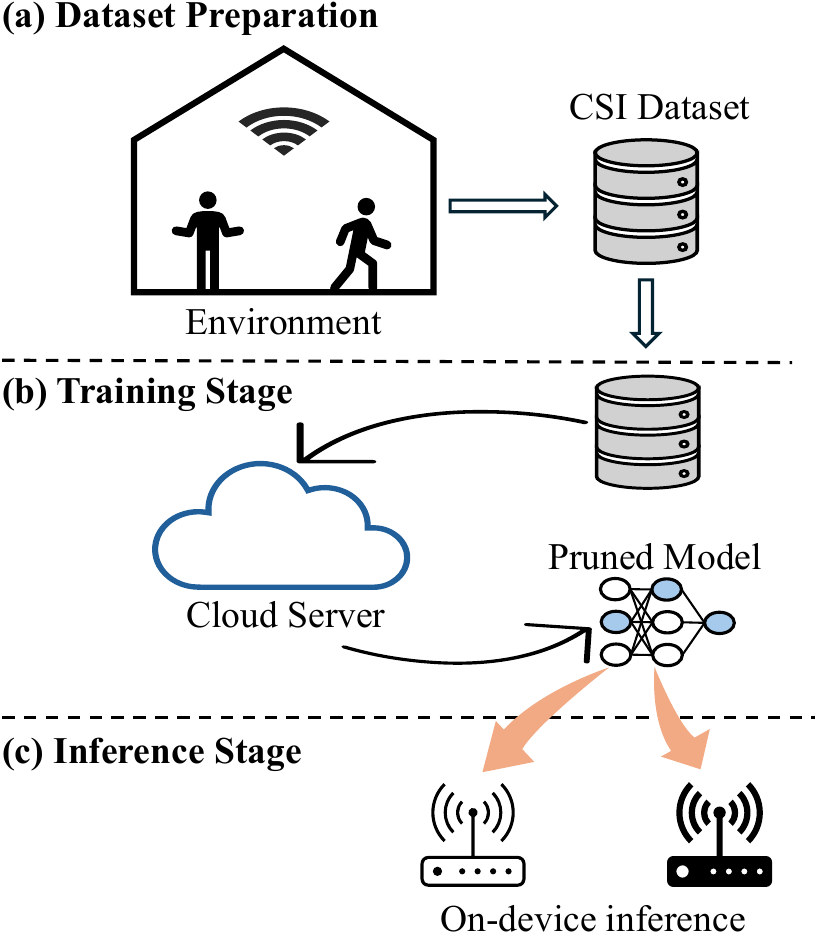}
\caption{Overview of the on-device Wi-Fi-based human activity recognition (HAR) system. 
(a) Dataset preparation, where CSI measurements are collected from human activities in an indoor Wi-Fi environment. 
(b) Training stage, where the CSI dataset is transferred to a cloud server for model training. 
(c) Inference stage, where the trained model is deployed on devices to perform real-time activity recognition.}
\label{HAR_system}
\end{figure}

In this study, we investigate a specific deployment scenario for Wi-Fi-based HAR system deployment on devices, where computational resources, energy availability, and memory are significantly constrained. As shown in Fig.~\ref{HAR_system}, an on-device Wi-Fi HAR system is typically structured into three primary stages: dataset preparation, training, and inference.

During the dataset preparation stage (Fig.~\ref{HAR_system}(a)), CSI data is collected in an indoor environment where participants perform a predefined set of activities. These human motions induce characteristic fluctuations in the wireless channel, which are captured as time-series CSI measurements. 

Once the collection is complete, the dataset is transferred to a cloud computing server  (Fig.~\ref{HAR_system}(b)) that provides computational resources for signal processing and model training. 

Since our objective is on-device deployment, the DL model undergoes compression techniques such as pruning to reduce its size before being deployed on the target devices. Finally, in the inference stage (Fig.~\ref{HAR_system}(c)), the compressed model is deployed onto the target devices (e.g., routers or IoT nodes) to classify human activities in real-time.
\subsection{Signal Processing}
The raw CSI data is a complex-valued tensor $H$. We use only the amplitude part $|H_k|$ as the model input, since the raw phase is often affected by hardware-induced sampling frequency offsets and carrier frequency offsets, which can reduce its reliability for robust HAR~\cite{YANG2023100703}. The two datasets come in different shapes, so we use a small dataset-specific initial block (Fig.~\ref{ResNet}(b)) and (Fig.~\ref{ResNet}(c)) to bring them into a common format before passing them through the shared ResNet-18 backbone:
\begin{itemize}
    \item \textbf{UT-HAR}: Following SenseFi~\cite{YANG2023100703}, each sample is reshaped into a single-channel 2D matrix $\mathbf{x} \in \mathbb{R}^{1 \times 250 \times 90}$, where $250$ is the number of time snapshots and $90$ comes from $3$ receive antennas $\times$ $30$ subcarriers. Each sample is then scaled to the range $[0, 1]$ using a simple per-sample min--max normalization: $\tilde{\mathbf{x}} = (\mathbf{x} - \min(\mathbf{x})) / (\max(\mathbf{x}) - \min(\mathbf{x}))$.
    \item \textbf{Widar3.0}: Following SenseFi~\cite{YANG2023100703}, each sample is a tensor $\mathbf{x} \in \mathbb{R}^{22 \times 20 \times 20}$, where $22$ is the time dimension and $20 \times 20$ is the velocity grid. Each sample is standardized using the dataset mean $\mu = 0.0025$ and standard deviation $\sigma = 0.0119$, computed once on the Widar3.0 training set: $\tilde{\mathbf{x}} = (\mathbf{x} - \mu) / \sigma$.
\end{itemize}
We use different normalization schemes for the two datasets because each one follows the convention of its original release~\cite{9516988,8067693}. Moreover, this signal processing pipeline is consistent with the SenseFi benchmark~\cite{YANG2023100703}, so our results can be directly compared with earlier HAR studies.
\subsection{Deep Learning Model}
\label{DL_Model}
While various DL architectures have been investigated for Wi-Fi-based HAR, including MLPs, CNNs, LSTM, Vision Transformers (ViTs), and Residual Networks (ResNets)~\cite{YANG2023100703}, recent studies indicate that ResNet-based architectures provide strong robustness and high classification accuracy~\cite{YANG2023100703}. In particular, benchmarking results across multiple public datasets show that ResNet-18 consistently achieves the highest classification accuracy on both the UT-HAR and Widar3.0 datasets~\cite{YANG2023100703}. Therefore, we adopt ResNet-18 as the representative model for Wi-Fi-based HAR and use it as the case study in this work.

\begin{figure}[!t]
\centering
\includegraphics[width=3.4in]{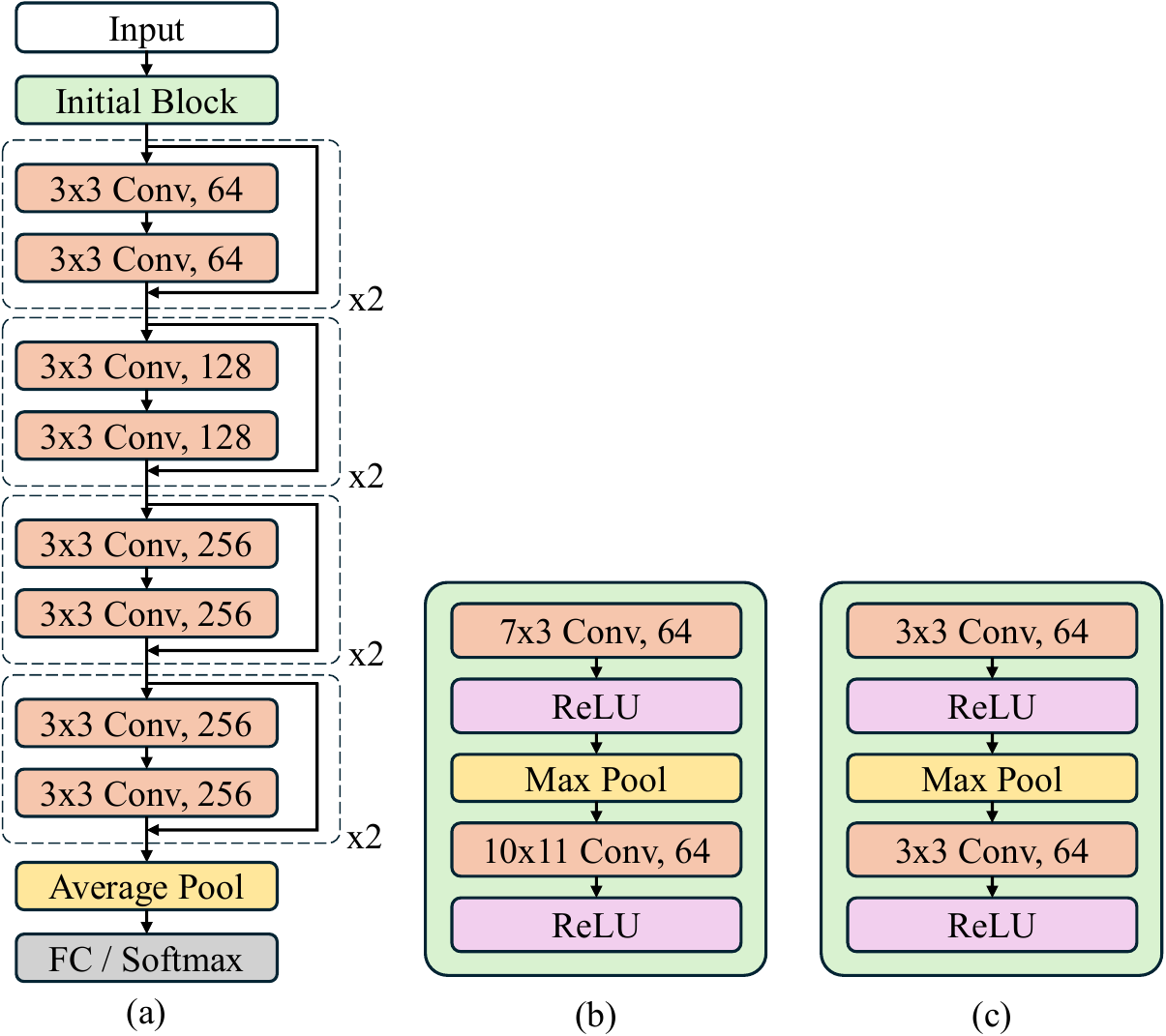}
    \caption{Architecture of the ResNet-18 model. (a) Overall architecture. (b) Initial block used in UT-HAR. (c) Initial block used in Widar3.0.}
\label{ResNet}
\end{figure}

The architecture of ResNet-18 is shown in Fig.~\ref{ResNet}. The ResNet-18 model has about 11.6 million trainable parameters, which enables strong classification performance, but also makes it relatively expensive to train and deploy.
To accommodate the distinct input tensor shapes of the UT-HAR and Widar3.0 datasets, specific initial blocks (as shown in Fig.~\ref{ResNet}(b) and Fig.~\ref{ResNet}(c)) are employed to unify the input dimensions. This enables the use of a common core network architecture across the two datasets.

\section{Quantum-Assisted Memory-Efficient Training Framework}
\subsection{Framework Overview}
\begin{figure*}[!t]
\centering
\includegraphics[width=6.5in]{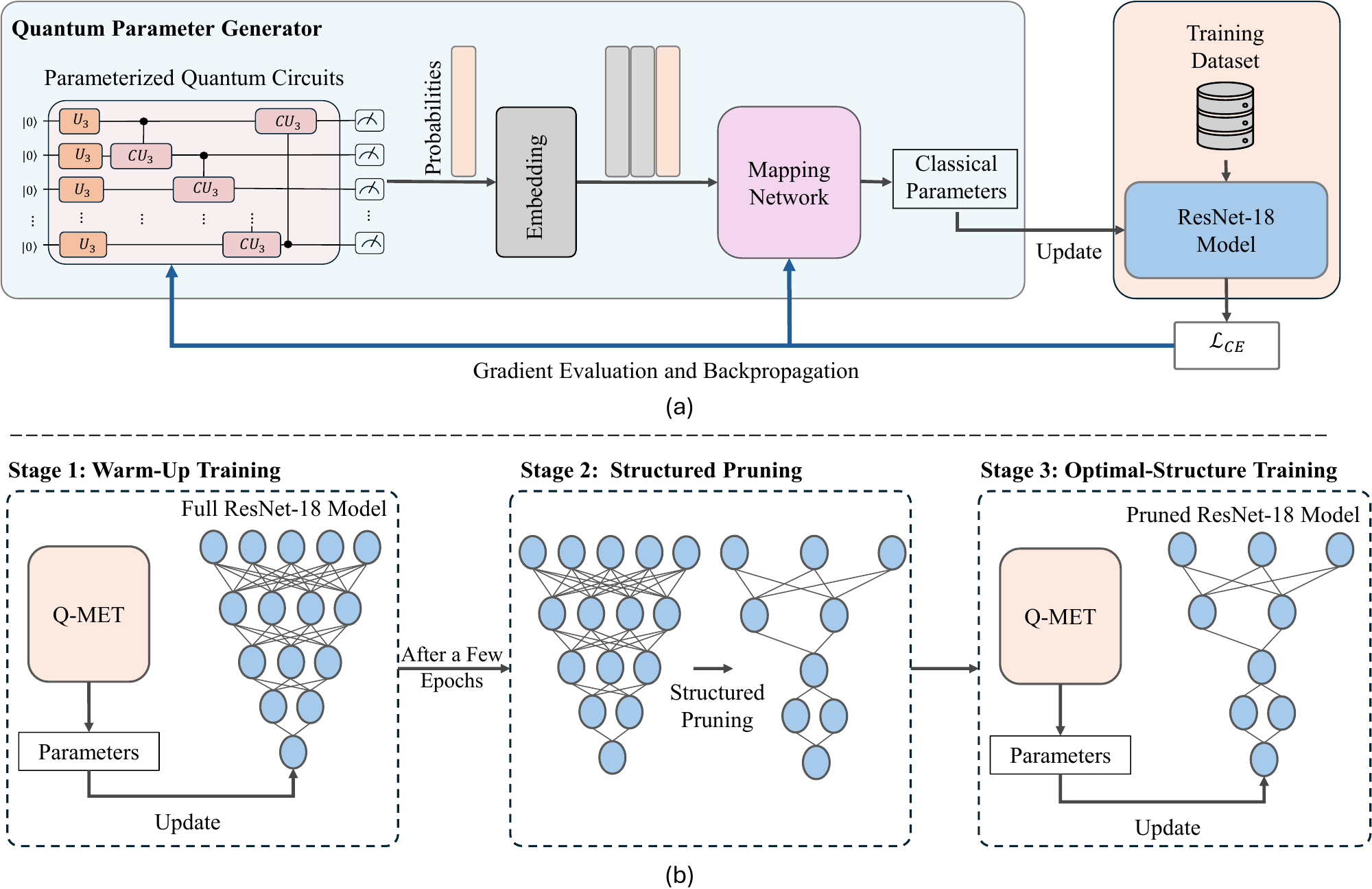}
    \caption{Proposed quantum-assisted memory-efficient training framework. (a) Overall architecture where the Q-MET generates parameters for ResNet-18. (b) Training workflow of Q-MET: warm-up training, LAMP-based pruning, and optimal-structure training.}
\label{Q_MET}
\end{figure*}

As shown in Fig.~\ref{Q_MET}(a), the Q-MET framework employs a quantum parameter generator (QPG) to dynamically generate the parameters for a classical ResNet-18 model. The QPG comprises three main components: a PQC, a sinusoidal embedding layer, and a classical mapping network.  

In the PQC, the $U_3$ gate is employed to manipulate qubit states, while the $CU_3$ gate establishes entanglement between qubits~\cite{10821103}. The $U_3$ and $CU_3$ gates  are defined as follows
\begin{equation}
U_3(\theta, \varphi, \lambda) = 
\begin{bmatrix}
\cos(\theta/2) & -e^{i\lambda} \sin(\theta/2) \\
e^{i\varphi} \sin(\theta/2) & e^{i(\varphi + \lambda)} \cos(\theta/2)
\end{bmatrix},
\end{equation}

% \begin{equation}
% CU_3 = I \otimes |0\rangle\langle 0| + U_3(\tilde{\theta},\tilde{\varphi},\tilde{\lambda}) \otimes |1\rangle\langle 1|,
% \end{equation}
\begin{equation}
\label{CU3}
CU_3(\tilde{\theta},\tilde{\varphi},\tilde{\lambda})_{q_0,q_1}
=\begin{bmatrix}
1 & 0 & 0 & 0 \\
0 & \cos\!\left(\tfrac{\tilde{\theta}}{2}\right) & 0 & -e^{i\tilde{\lambda}}\sin\!\left(\tfrac{\tilde{\theta}}{2}\right) \\
0 & 0 & 1 & 0 \\
0 & e^{i\tilde{\varphi}}\sin\!\left(\tfrac{\tilde{\theta}}{2}\right) & 0 & e^{i(\tilde{\varphi}+\tilde{\lambda})}\cos\!\left(\tfrac{\tilde{\theta}}{2}\right)
\end{bmatrix},
\end{equation}

\noindent where $(\theta, \varphi, \lambda)$ and $(\tilde{\theta}, \tilde{\varphi}, \tilde{\lambda})$ are real-valued parameters that define the rotations applied by the $U_3$ and $CU_3$ gates, $q_0$ and $q_1$ denote the control and target qubits, respectively.

The $CU_3$ is designed with a circular layout, which enables for constructing multi-qubits interactions within the PQC. The PQC can be expressed as
\begin{equation}
    |\psi(\theta_{PQC})\rangle = U(\theta_{PQC})|0\rangle^{\otimes N_q},
\end{equation}
where $N_q$ denotes the number of qubits, and total number of trainable parameters in the PQC is $\theta_{PQC} =  6 N_q$, as each $U_3$ and $CU_3$ gate has three tunable parameters. 
The measurement probabilities of the PQC output can be expressed as a probability vector as follows
\begin{equation}
\Psi = 
\begin{bmatrix}
|\langle \phi_1 | \psi(\theta_{PQC}) \rangle|^2 \\
\vdots \\
|\langle \phi_k | \psi(\theta_{PQC}) \rangle|^2 \\
\vdots \\
|\langle \phi_{\mathcal{P}} | \psi(\theta_{PQC}) \rangle|^2
\end{bmatrix},
\end{equation}
\noindent
where $|\phi_k\rangle \in \left\{0,1  \right\}^{N_q}$ represents the $k$-th computational basis state, 
$k \in \{1, 2, \ldots, \mathcal{P}=2^{N_q}\}$ indexes the computational basis state, and 
$\left| \langle \phi_k | \psi(\theta_{PQC}) \rangle \right|^2$ 
is the probability of obtaining the outcome 
$|\phi_k\rangle$ upon measurement.

These probabilities are then further processed by an embedding function that adds information to represent the uniqueness of computational basis state. 
In this study, we employ sinusoidal embedding, as proposed in~\cite{11575585}, because it offers a compact and uniquely identifiable representation of computational basis states, while significantly reducing the embedding dimensionality from exponential to constant size, thereby improving scalability and computational efficiency.
The sinusoidal embedding is defined as follows
\begin{equation}
\label{proposed_embedding_formula}
\mathcal{S}(\Psi)=  \left[ S_0,S_1, \Psi \right],
\end{equation}
\noindent where $S_0$ and $S_1$ represent embedding vectors that serve as compact and distinctive identifiers for each set of measurement probabilities. As a result, the triplet $\left[ S_0,S_1, \Psi \right]$ effectively encodes the identity of the corresponding computational basis state linked to each probability. The explicit expansion of
Eq. (\ref{proposed_embedding_formula}) is presented as follows
\begin{equation}
\mathcal{S}(\Psi)= 
\left[
\begin{array}{ccc}
S_{0, 0} , &S_{1, 0} ,&   |\langle \phi_1 | \psi(\theta_{PQC}) \rangle|^2 \\
\vdots & \vdots \\
S_{0, k_e} , &S_{1, k_e} ,& |\langle \phi_k | \psi(\theta_{PQC}) \rangle|^2 \\
\vdots & \vdots \\
S_{0, \mathcal{P}-1 } , &S_{1, \mathcal{P}-1 } ,&   |\langle \phi_ \mathcal{P} | \psi(\theta_{PQC}) \rangle|^2 \\
\end{array}
\right],
\end{equation}
where $S_{0, k_e}$ and $S_{1, k_e}$ are the two sinusoidal embedding components, defined as
\begin{equation}
    S_{\iota, k_e} = \sin\left( \frac{k_e}{\mathcal{P}} 2\pi + \frac{\iota\pi}{2} \right),
\end{equation}
\noindent
where $\iota = 0, 1$ is the embedding index, and $k_e$ is the quantum state index with $k_e \in \{0, 1, 2, \ldots, \mathcal{P}-1\}$.

% \begin{table}[t]
% \centering
% \caption{Detailed architecture of the Mapping Network model}
% \label{mapping_network_architecture}
% \begin{tabular}{llC{1.2cm}c}
% \toprule
% \textbf{Component} & \textbf{Layer Type} & \textbf{Output Shape} & \textbf{Parameters} \\
% \midrule
% \textbf{Inner Function}$^{\dagger}$ 
%     & FC (1 → 16) + BN  & $(n_G,16)$  & 64 \\
%     & FC (6 → 1)       & $(n_G,1)$  & 17 \\
% \midrule
% \textbf{Outer Function} 
%     & FC (3 → 32) + BN  & $(n_G,32)$  & 192 \\
%     & FC (32 → 64)      & $(n_G,64)$ & 2112 \\
%     & FC (64 → 5)      & $(n_G,5)$  & 325 \\
%     & FC (5 → $n_B$)   & $(n_G,n_B)$ & $6 \times n_B$ \\
% \midrule
% \textbf{Total} & - & - & $2872 + 6 \times n_B$ \\
% \bottomrule
% \end{tabular}%
% \\[1ex]
% \footnotesize\textit{$^{\dagger}$The inner function is applied independently to each of the 3 input variables (repeated 3 times)}
% \end{table}
\begin{figure}[!t]
\centering
\includegraphics[width=3.4in]{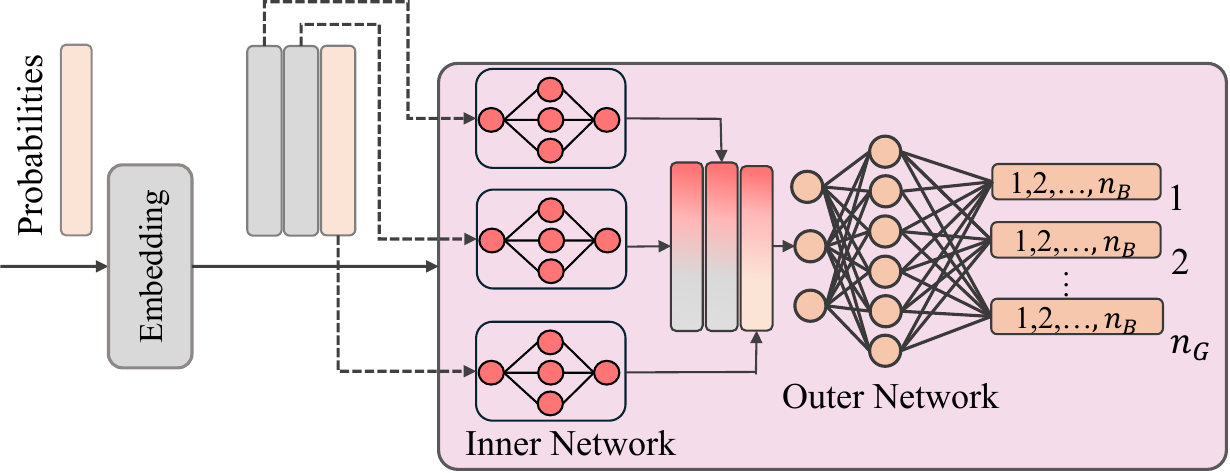}
    \caption{Overview of mapping network in the proposed quantum-assisted memory-efficient training framework.}
\label{Mapping_Network}
\end{figure}
\begin{table}[t]
\centering
\caption{Detailed architecture of mapping network model}
\label{mapping_network_architecture}
\begin{tabular}{llc}
\toprule
\textbf{Component} & \textbf{Layer Type} & \textbf{Parameters} \\
\midrule
Inner Function$^{\dagger}$ 
    & FC (1 → 16) + BN  & 64 \\
    & FC (16 → 1)       & 17 \\
\midrule
Outer Function 
    & FC (3 → 32) + BN & 192 \\
    & FC (32 → 64)     & 2112 \\
    & FC (64 → 5)      & 325 \\
    & FC (5 → $n_B$)   & $6 \times n_B$ \\
\midrule
Total & - & $2872 + 6 \times n_B$ \\
\bottomrule
\end{tabular}%
\\[1ex]
\footnotesize\textit{$^{\dagger}$The inner function is applied independently to each of the 3 input variables (repeated 3 times).}
\end{table}

In the Q-MET framework, these embedded probability vectors are then mapped to the parameters of the ResNet-18 model using a classical neural network, referred to as the mapping network, denoted by $M$ with tunable parameters $\theta_m$. As depicted in Fig.~\ref{Mapping_Network}, the mapping network comprises an inner network and an outer network. Detailed layer specifications are provided in Table~\ref{mapping_network_architecture}. The structure of the mapping network follows the sinusoidal-embedding mapping design introduced in our prior QT work~\cite{11575585}, where an inner function is applied independently to each input variable and is followed by an outer function. Building on that design, the intermediate layer widths in Table~\ref{mapping_network_architecture} were enlarged and selected through a small search over candidate dimensions, guided by the trade-off between classification accuracy and the size of the mapping network. Since the mapping network accounts for the dominant share of the trainable parameters in the Q-MET framework, the smallest widths that retained competitive accuracy were chosen to keep the overall trainable-parameter count as low as possible. The output width $n_B$ is dataset-dependent and is set by~\eqref{eq:N_B} so that the total number of generated parameters matches the size of the target model, while the remaining widths are fixed across both datasets. The output of mapping network is a matrix parameter, which can be modeled as

\begin{equation}
     \boldsymbol{\Theta}_m=M_{\theta_m}(\mathcal{S}(\Psi)),
\end{equation}
where $\boldsymbol{\Theta}_m \in \mathbb{R}^{n_G \cdot n_B}$ denotes the matrix of classical parameters, with $n_G$ and $n_B$ representing the number of parameter groups and the number of parameters in each group, respectively. 

The output matrix is then reshaped into a 1D vector of the length $n_G \cdot n_B$, denoted as $\overrightarrow\theta_{ResNet-18}$, which serves as the classical parameter set of ResNet-18. It is worth noting that the QPG acts purely as a weight generator and does not take CSI data as input. This makes Q-MET applicable to any classical DL architecture regardless of the input modality.

\subsection{Structured Pruning}
To address inference inefficiency, Q-MET employs a structured channel pruning strategy guided by layer-adaptive magnitude-based pruning (LAMP)~\cite{lee2021layeradaptive} scores, where channel importance is derived from the LAMP scores of their associated weights. Compared to uniform or global magnitude pruning, LAMP better preserves sensitive layers, resulting in improved accuracy retention under high sparsity, while maintaining significant reductions in model size and computational cost. It is worth noting that Q-MET is compatible with any pruning technique, from unstructured to structured methods. Structured pruning is preferred here because it produces regular, dense matrices that are well-suited to hardware acceleration on resource-constrained platforms, whereas unstructured pruning yields irregular sparse matrices that are ineffectual for practical inference speedup~\cite{LIANG2021370}. Among structured criteria, LAMP is adopted for its layer-adaptive importance scoring, which better preserves sensitive layers under high sparsity compared to uniform magnitude pruning~\cite{lee2021layeradaptive}.

Consider a neural network in which each layer is associated with a weight tensor corresponding to a fully connected or convolutional layer. To provide a unified formulation across different layer types, each weight tensor $W$ is reshaped into a one-dimensional vector (i.e., vectorized). Without loss of generality, the elements of this vector are sorted and indexed in ascending order of magnitude, such that
\begin{equation}
|W[u]| \le |W[v]| \quad \text{for } u < v,
\end{equation}
where $W[u]$ denotes the $u$-th element of the flattened weight tensor $W$, and $u$ and $v$ index the entries of this vector after sorting in ascending order of magnitude.

Based on this ordering, the LAMP score assigned to the weight indexed by $u$ is defined as~\cite{lee2021layeradaptive}
\begin{equation}
\text{score}(u; W) = \frac{(W[u])^{2}}{\sum_{v \ge u} (W[v])^{2}}.
\end{equation}

This score measures the relative importance of a weight with respect to the remaining higher-magnitude weights within the same layer, as the denominator represents the cumulative squared magnitude of all weights ranked at or above index $u$.

To extend LAMP to structured pruning, a channel-level importance score is computed by aggregating the LAMP scores of all weights within each channel. After computing the channel scores, all channel scores across the network are concatenated into a single global score vector. Given a target pruning ratio $p$, channels with the smallest scores are globally removed until the desired sparsity level is reached.

\subsection{Training Flow for Quantum-Assisted Memory-Efficient Training Framework}
As shown in Fig.~\ref{Q_MET}(b), the training process of Q-MET consists of three stages: warm-up training, structured pruning, and optimal-structure training.

In the warm-up training stage, the full model is trained for a few epochs. The goal is not convergence, but to establish a weight distribution sufficient for identifying redundant connections. 
This is followed by structured pruning using LAMP. The LAMP scores are evaluated to identify and remove less important filters, resulting in a sparse model architecture by keeping only the most informative components.
Finally, the pruned model, now considered a compact structure, is further trained using Q-MET, which produces parameters only for the active channels. Therefore, the number of parameters generated is significantly lower compared to the full model, resulting in higher memory efficiency during training.

\subsubsection{Quantum Parameter Generator Procedure}
To construct the QPG, we first determine the number of qubits required to construct the PQC.
Based on the number of qubits and the total number of parameters in the classical ResNet-18 model, we then design the mapping network by specifying the parameters $n_B$ and $n_G$, which are given by
\begin{equation}
\label{eq:N_B}
n_G = 2^{N_q}, \quad n_B = \left\lceil \frac{\mathcal{C}_{\text{ResNet-18}}}{2^{N_q}} \right\rceil.
\end{equation}

Once the QPG is initialized, it is used to dynamically generate the parameters of the ResNet-18 model during training. At each training iteration, the QPG produces a parameter set $\overrightarrow{\theta}_{\text{ResNet-18}}$ corresponding only to the active (unpruned) components of the network. These parameters are then injected into the ResNet-18 architecture using PyTorch’s functional operators (e.g., \texttt{F.conv2d}, \texttt{F.linear}), which accept \texttt{weight} and \texttt{bias} as arguments and thus enable on-the-fly parameter injection into the corresponding layers. After the parameter update, the model processes a training input $x$ and produces a prediction $\hat{y}$ via the forward pass
\begin{equation}
    \hat{y} = Z(x; \overrightarrow\theta_{ResNet-18}),
\end{equation}
where  $Z(\cdot)$  denotes the forward function of ResNet-18 with the parameters generated by Q-MET.

\subsubsection{Loss Function and Gradient Estimation}
For multi-class classification tasks, the training loss is computed using the cross-entropy (CE) loss function

\begin{equation}
\label{CE_loss}
\mathcal{L}_{\text{CE}} = -\frac{1}{N_{\text{Batch}}} \sum_{i=1}^{N_{\text{Batch}}} \sum_{c=1}^{N_c} y_{i,c} \log(\hat{y}_{i,c}),
\end{equation}

\noindent where  $y_{i,c}$  denotes the ground-truth label of the  $i$-th sample for class $c$, and  $\hat{y}_{i,c}$ represents the predicted probability assigned to class $c$ by the model. $N_{\text{Batch}}$ is the batch size and $N_c $ denotes the number of output classes.

Since the parameters of ResNet-18 are generated by QPG, which consists of both the PQC with parameters $\theta_{PQC}$ and the classical mapping network with parameters $\theta_{m}$, the loss gradient needs to be propagated through the entire QPG pipeline~\cite{10821200}. By applying the chain rule, the gradient of the loss with respect to the joint QPG parameters is computed as~\cite{10821200}
\begin{equation}
\nabla_{(\theta_{\text{PQC}}, \theta_{m})} \mathcal{L}_{\text{CE}} = 
\left( \frac{\partial \theta_{\text{ResNet-18}}}{\partial (\theta_{\text{PQC}}, \theta_{m})} \right)^\top 
\nabla_{\theta_{\text{ResNet-18}}} \mathcal{L}_{\text{CE}},
\end{equation}
where $\partial \theta_{\text{ResNet-18}}/\partial (\theta_{\text{PQC}}, \theta_{m})$ denotes the Jacobian matrix, which quantifies the sensitivity of the generated ResNet-18 parameters with respect to the PQC and mapping network parameters.

The parameters of the PQC and the mapping network are updated via gradient descent as follows
\begin{equation}
\theta_{PQC}^{(t+1)},\ \theta_{m}^{(t+1)} = \theta_{PQC}^{(t)},\ \theta_{m}^{(t)} - \eta \nabla_{(\theta_{PQC}, \theta_{m})} \mathcal{L}_{\text{CE}},
\end{equation}
where $\eta$ represents the learning rate. 
\subsection{Parameter Efficiency Analysis}
\label{subsec:efficiency_analysis}
During training, the Q-MET framework dynamically generates the parameters required for the ResNet-18 model, thereby eliminating the need to train and update the massive number of parameters in ResNet-18. Instead, the trainable parameters are restricted to the compact QPG, which significantly reduces memory overhead and enables efficient training in resource-constrained environments. This reduction is particularly important, as memory consumption during training arises not only from model weights, but also from optimizer states, activation tensors, and gradient buffers. By minimizing the trainable parameter count, Q-MET significantly lowers the memory overhead for gradients and optimizer states, enabling efficient training in resource-constrained environments.

This subsection quantifies the efficiency of Q-MET in terms of the number of trainable parameters. Let $\mathcal{C}_{\text{ResNet-18}}$  denote the total number of parameters in the baseline ResNet-18 model. In contrast, the parameter count of Q-MET consists of the combined contributions from the PQC and the mapping network, which can be modeled as

\begin{equation}
C_{\text{Q-MET}} = \underbrace{6N_q}_{\text{PQC}} + \underbrace{(2872 + 6n_B)}_{\text{Mapping Network}},
\end{equation}

When substituting (\ref{eq:N_B}), the following results are obtained

\begin{equation}
\label{eq:qmet_param}
\mathcal{C}_{\text{Q-MET}} = 6N_q+ \left\lceil \frac{6\mathcal{C}_{\text{ResNet-18}}}{2^{N_q}} \right\rceil + 2872.
\end{equation}

The parameter efficiency gain is then defined as

\begin{equation}
\label{C_percentage}
\Delta \mathcal{C}(\%) = \left(1 - \frac{\mathcal{C}_{\text{Q-MET}}}{\mathcal{C}_{\text{ResNet-18}}}\right) \times 100,
\end{equation}
Substituting (\ref{eq:qmet_param}) yields
\begin{equation}
\label{eq:qmet_eff}
\Delta \mathcal{C}(\%) =
\left(
1 - \frac{6N_q}{\mathcal{C}_{\text{ResNet-18}}}
      - \frac{6}{2^{N_q}}
      - \frac{2872}{\mathcal{C}_{\text{ResNet-18}}}
\right) \times 100.
\end{equation}

In the Q-MET framework, $N_q$ is significantly smaller than $\mathcal{C}_{\text{ResNet-18}}$ (approximately $11.6 \times 10^6$), both terms $6N_q / \mathcal{C}_{\text{ResNet-18}}$ and $2872 / \mathcal{C}_{\text{ResNet-18}}$ become negligible. Therefore,~(\ref{eq:qmet_eff}) can be approximated as

\begin{equation}
\label{qmet_eff_approx}
\Delta \mathcal{C}_{\text{approx}}(\%) \approx \left(1 - \frac{6}{2^{N_q}}\right) \times 100.
\end{equation}

From (\ref{qmet_eff_approx}), the relationship between the number of qubits and the corresponding parameter efficiency gain is summarized in Table~\ref{tab:qmet_efficiency}. When $N_q = 2$, the negative gain ($-50\%$) indicates that Q-MET requires more parameters than the baseline ResNet-18 model, and is therefore less efficient at very low qubit counts, but begins to outperform the baseline when $N_q \geq 3$, where the gain is approximately $25\%$.

\begin{table}[t]
\caption{Approximate Parameter Efficiency Gains Achieved by Q-MET Across Different Qubit Counts}
\label{tab:qmet_efficiency}
\centering
\renewcommand{\arraystretch}{1.15}
\setlength{\tabcolsep}{8pt}
\begin{tabular}{c c}
\hline
\textbf{Number of Qubits ($N_q$)} & \textbf{$\Delta \mathcal{C}_{\text{approx}}(\%)$} \\
\hline
2  & $-50.00$ \\
3  & $25.00$  \\
4  & $62.50$  \\
5  & $81.25$  \\
6  & $90.63$  \\
7  & $95.31$  \\
8  & $97.66$  \\
9  & $98.83$  \\
10 & $99.41$  \\
\hline
\end{tabular}
\end{table}
As $N_q$ increases, the efficiency improves rapidly. When $N_q = 4$, the gain rises to 
$62.5\%$, increasing to $81.25\%$ at $N_q = 5$, reaching $90.63\%$ at $N_q = 6$. These results demonstrate that more than 90\% parameter efficiency can be achieved using only six qubits. This highlights the potential of Q-MET to reduce model complexity in a highly efficient manner with relatively low quantum resource requirements.

Beyond six qubits, however, the additional gains become smaller. For example, increasing $N_q$ from $6$ to $10$ yields an additional gain of only $8.78\%$, reaching a maximum of $99.41\%$. This suggests a point of diminishing returns, where most of the parameter savings are achieved early, and further increases in qubit count provide limited improvements. 

\subsection{Computational Complexity Analysis}

The per-iteration training complexity of Q-MET consists of three main components: the PQC, the mapping network, and the pruned ResNet-18. When the PQC is simulated classically and the full probability vector is evaluated, its cost scales as $\mathcal{O}(2^{N_q})$. The mapping network generates $n_G n_B$ parameters and therefore has complexity $\mathcal{O}(n_G n_B)$. For the pruned ResNet-18, let $F_{\mathrm{ResNet\text{-}18}}$ denote the training cost of the full model. At sparsity level $s$, the cost is approximated as $\mathcal{O}((1-s)F_{\mathrm{ResNet\text{-}18}})$.
Thus, the total per-iteration complexity of Q-MET is
\begin{equation}
    \mathcal{O}_{\mathrm{Q\text{-}MET}}
    =
    \mathcal{O}(2^{N_q})
    +
    \mathcal{O}(n_G n_B)
    +
    \mathcal{O}((1-s)F_{\mathrm{ResNet\text{-}18}}).
\end{equation}

In comparison, standard backpropagation-based training of the full ResNet-18 has complexity $\mathcal{O}(F_{\mathrm{ResNet\text{-}18}})$. Therefore, Q-MET introduces additional parameter-generation overhead, which is consistent with the empirical training-time increase of 18\%--36\% observed in Fig.~5.

The main benefit of Q-MET is memory efficiency rather than reduced computation. Standard training stores weights, gradients, and optimizer states for all ResNet-18 parameters, resulting in model-state memory scaling with $\mathcal{O}(C_{\mathrm{ResNet\text{-}18}})$. In contrast, Q-MET maintains trainable states only for the compact QPG, reducing this component to $\mathcal{O}(C_{\mathrm{Q\text{-}MET}})$, where $C_{\mathrm{Q\text{-}MET}} \ll C_{\mathrm{ResNet\text{-}18}}$.
\section{Experimental Evaluation}
\subsection{Dataset}

\begin{table}[]
\centering
\caption{Summary of Datasets}
\label{Summary_of_Datasets}
% \resizebox{0.95\columnwidth}{!}{%
\begin{tabular}{|c|c|c|}
\hline
\textbf{Datasets}              & \textbf{UT-HAR}         & \textbf{Widar3.0}       \\ \hline
Hardware Platform     & Intel 5300 NIC & Intel 5300 NIC \\ \hline
Number of Classes     & 7              & 22             \\ \hline
Sample Dimensionality & (1,250,90)      & (22,20,20)     \\ \hline
Training samples      & 3,977          & 34,926         \\ \hline
Testing samples       & 996            & 8,726          \\ \hline
\end{tabular}%
% }
\end{table}
The experiments were conducted using two public datasets, which are UT-HAR~\cite{8067693} and Widar3.0~\cite{9516988}. Table~\ref{Summary_of_Datasets} summarizes their specifications. Both datasets contain CSI data collected using the Intel 5300 Wi-Fi network interface card under real-world conditions. 

The UT-HAR dataset contains 7 different human motions, namely \textit{lie down, fall, walk, pick up, run, sit down, stand up}. These activities cover a range of basic human motions and are commonly used to evaluate the performance of CSI-based HAR models~\cite{8067693, YANG2023100703}. 

In contrast, the Widar3.0 dataset is a large-scale benchmark focused on fine-grained gesture recognition. These classes include common gesture primitives such as \textit{push \& pull, sweep, clap, and slide}, as well as 18 drawing gestures~\cite{9516988,YANG2023100703}. The Widar3.0 dataset is widely used to assess the scalability, discriminative capabilities, and cross-domain generalization of CSI-based sensing systems~\cite{YANG2023100703}.
\subsection{Training Hyperparameter Configuration}
Training was conducted on a Dell Precision 3680 workstation equipped with an Intel\textsuperscript{\textregistered} Core\texttrademark{} i7-14700 processor, 64~GB of RAM, a 2~TB SSD, and an NVIDIA GeForce RTX 4090 GPU. We used the Adam optimizer with an initial learning rate of 0.001 and a batch size of $128$. Training ran for up to $500$ epochs, with an early stopping mechanism triggered if validation loss did not improve for $30$ consecutive epochs.
\subsection{Evaluation Metrics}
\subsubsection{Accuracy}
Accuracy is a key metric for assessing the classification performance of DL-based HAR models. It represents the ratio of correctly predicted instances to the total number of instances in the test set. The accuracy can be expressed as
 \begin{equation}
Acc = \frac{\text{Number of Correct Predictions}}{\text{Total Number of Predictions}}.
\end{equation}

To evaluate the impact of Q-MET relative to the traditional approach, we define the accuracy loss as
\begin{equation}
    Acc_{Loss} = Acc_{baseline} - Acc_{\text{Q-MET}} , 
\end{equation}
\noindent where $Acc_{\text{Q-MET}}$ and $Acc_{baseline}$ denote the accuracies of the ResNet-18 model trained using the Q-MET framework and the classical approach, respectively.

\subsubsection{Parameter Efficiency}
\label{subsec:parameter_efficiency}

We use the number of trainable parameters as a primary metric to evaluate efficiency, as parameter count is the canonical and widely accepted efficiency measure in the parameter-efficient training literature such as LoRA~\cite{hu2022lora}, and the QT family~\cite{10821200,11575585,liu2025a,10821046,10821103}.

For a DL model with $\mathcal{C}$ trainable parameters, the model-state component of training memory, comprising the parameters, gradients, and the two Adam moment estimates stored in FP32, can be expressed as~\cite{3433727}
\begin{equation}
\label{eq_training_mem}
M_{\text{model}}^{\text{train}} = M_p + M_{\text{grads}} + M_{\text{opt}},
\end{equation}
where $M_p = \mathcal{C}\mathcal{S}$ denotes the memory required to store the model parameters, $M_{\text{grads}} = \mathcal{C}\mathcal{S}$ corresponds to the memory for storing gradients, and $M_{\text{opt}} = 2\mathcal{C}\mathcal{S}$ represents the memory required by the optimizer states when using the Adam optimizer. Here, $\mathcal{S}$ denotes the number of bytes per parameter (e.g., $\mathcal{S}=4$ for FP32).

Therefore,~(\ref{eq_training_mem}) can be rewritten as
\begin{equation}
\label{eq_training_mem_simple}
M_{\text{model}}^{\text{train}} = 4\mathcal{C}\mathcal{S}.
\end{equation}

Under FP32 precision, this corresponds to $M_{\text{model}}^{\text{train}} = 16\mathcal{C}$ bytes. Concretely, the model-state memory of the baseline ResNet-18 is approximately $177$~MB on UT-HAR and $171$~MB on Widar3.0, whereas Q-MET reduces it to about $8.4$~MB (QT-7) and $17$~MB (QT-6), respectively, corresponding to a reduction of roughly $95\%$ and $90\%$. These figures follow directly from the trainable-parameter counts and the relation $M_{\text{model}}^{\text{train}} = 16\mathcal{C}$. We emphasize that this quantifies the model-state component of training memory (parameters, gradients, and optimizer states). The activation and buffer memory, which is determined by the network forward pass and batch size, is orthogonal to and not the target of the proposed framework, though structured pruning incidentally reduces it.

During inference, gradient and optimizer states are not required, and the memory footprint is primarily determined by the model parameters. As a result, inference memory remains strongly correlated with the parameter count, further justifying the use of parameter efficiency as a unifying metric for evaluating both training and inference memory requirements.

To quantify the efficiency of the proposed Q-MET framework, we adopt the notion of parameter efficiency, as defined in (\ref{C_percentage}). This metric measures the percentage reduction in the number of trainable parameters achieved by Q-MET relative to the classical training approach.

\subsubsection{Sparsity}
Sparsity is a key metric for assessing the effectiveness of structured pruning~\cite{0366}. It represents the percentage of the model that has been pruned and can be defined as

\begin{equation}
\text{Sparsity} (\%)= \frac{\text{Number of pruned parameters}}{\text{Total number of parameters}} \times 100.
\end{equation}
\noindent A higher sparsity level indicates a more aggressively pruned model, which typically results in reduced model size and faster inference times~\cite{0366}.
\subsubsection{Training Time}
To evaluate the computational efficiency of the QT framework, we define the average training time per epoch, $T_e$ (in seconds), as
\begin{equation}
T_e = \frac{\text{Total training time}}{\text{Number of epochs}},
\end{equation}
where the total training time is measured until model convergence.

\subsection{Experimental Results}
In this subsection, we first evaluate the effectiveness of the QT framework in training the classical ResNet-18 model to determine the optimal QT configuration. We then systematically analyze the performance of the proposed Q-MET framework under various training settings, including different numbers of Warm-Up epochs and pruning ratios. Finally, we discuss the observed performance-efficiency trade-offs and provide insights of Q-MET for resource-constrained HAR deployments.
\subsubsection{Performance of Classical ResNet-18 Model and QT Framework}
\label{Compare_classical_vs_QT_Time_complex}
We first examine the performance of the ResNet-18 model that is trained via conventional training method for both UT-HAR and Widar3.0 datasets, which serves as the baseline for our study. 
Table~\ref{Baseline_performance} reports the number of parameters, highest and lowest accuracy across five runs, average accuracy with standard deviation.

\begin{table}[t]
\centering
\caption{Baseline performance of ResNet-18 on UT-HAR and Widar3.0}
\label{Baseline_performance}
\resizebox{0.95\columnwidth}{!}{%
\begin{tabular}{|c|c|c|c|c|c|}
\hline
\multirow{2}{*}{\textbf{Model}} &
  \multirow{2}{*}{\textbf{Datasets}} &
  \multirow{2}{*}{\textbf{Parameters}} &
  \multicolumn{3}{c|}{\textbf{Acc (\%)}} \\ \cline{4-6}
 &          &            & \textbf{Highest} & \textbf{Lowest} & \textbf{Average} \\ \hline
\multirow{2}{*}{ResNet-18} &
  UT-HAR &
  11,622,599 &
  98.80 &
  97.60 &
  98.08~$\pm$~0.44 \\ \cline{2-6}
 & Widar3.0 &
  11,227,798 &
  72.22 &
  69.71 &
  71.29~$\pm$~1.02 \\ \hline
\end{tabular}%
}
\end{table}
ResNet-18 achieves strong classification performance on the UT-HAR dataset, yielding an average classification accuracy of 98.08 $\pm$ 0.44\%. In contrast, performance on the Widar3.0 dataset is lower, with an average accuracy of 71.29 $\pm$ 1.02\%. These results are consistent with those reported in~\cite{YANG2023100703}. The lower classification accuracy on the Widar3.0 dataset is primarily attributed to the greater complexity of the dataset, as the Widar3.0 dataset includes 22 gesture classes, compared to only 7 activity classes in the UT-HAR dataset. The larger number of classes increases the difficulty of the classification task, making it more challenging for the model to distinguish between similar gestures. Additionally, the number of parameters is slightly different between the two settings, with 11.6 million in UT-HAR and 11.2 million in Widar3.0. This difference is due to small adjustments in the initial block made for each dataset, as described in Section~\ref{DL_Model}.

Subsequently, we evaluate the effectiveness of the proposed QT framework for training the classical ResNet-18 models. As presented in Section IV-D, the QT framework starts attaining efficiency when the qubits count is more than three. To analyze the performance of the framework under various levels of compression, we conducted experiment with qubit settings ranging from 3 to 10 qubits (QT-3 to QT-10). As shown in Table~\ref{QT_full_comparison}, the actual parameter efficiency achieved by the QT framework ($\Delta \mathcal{C}(\%)$) closely aligns with the approximated values ($\Delta \mathcal{C}_{\text{approx}}(\%)$), thereby validating the theoretical analysis presented in Section~\ref{subsec:efficiency_analysis}.

\begin{table*}[]
\centering
\caption{Performance of ResNet-18 trained with QT using different numbers of qubits}
\label{QT_full_comparison}
% \resizebox{0.95\columnwidth}{!}{%
\begin{tabular}{|c|c|c|ccc|ccc|}
\hline
\multirow{3}{*}{\textbf{QT Variants}} &
  \multirow{3}{*}{\textbf{$\Delta \mathcal{C}_{\text{approx}}(\%)$}} &
  \multicolumn{1}{l|}{\multirow{3}{*}{\textbf{$\Delta \mathcal{C}(\%)$}}} &
  \multicolumn{3}{c|}{\textbf{UT-HAR}} &
  \multicolumn{3}{c|}{\textbf{Widar3.0}} \\ \cline{4-9} 
 &
   &
  \multicolumn{1}{l|}{} &
  \multicolumn{3}{c|}{\textbf{Acc (\%)}} &
  \multicolumn{3}{c|}{\textbf{Acc (\%)}} \\ \cline{4-9} 
 &
   &
  \multicolumn{1}{l|}{} &
  \multicolumn{1}{l|}{\textbf{Highest}} &
  \multicolumn{1}{l|}{\textbf{Lowest}} &
  \multicolumn{1}{l|}{\textbf{Average}} &
  \multicolumn{1}{l|}{\textbf{Highest}} &
  \multicolumn{1}{l|}{\textbf{Lowest}} &
  \multicolumn{1}{l|}{\textbf{Average}} \\ \hline
QT-3 &
  25.00 &
  24.98 &
  \multicolumn{1}{c|}{99.20} &
  \multicolumn{1}{c|}{99.00} &
  99.12~$\pm$~0.10 &
  \multicolumn{1}{c|}{73.02} &
  \multicolumn{1}{c|}{72.43} &
  72.78~$\pm$~0.22 \\ \hline
QT-4 &
  62.50 &
  62.48 &
  \multicolumn{1}{c|}{99.20} &
  \multicolumn{1}{c|}{98.80} &
  99.08~$\pm$~0.16 &
  \multicolumn{1}{c|}{73.14} &
  \multicolumn{1}{c|}{71.82} &
  72.35~$\pm$~0.47 \\ \hline
QT-5 &
  81.25 &
  81.22 &
  \multicolumn{1}{c|}{99.40} &
  \multicolumn{1}{c|}{98.80} &
  99.20~$\pm$~0.22 &
  \multicolumn{1}{c|}{71.74} &
  \multicolumn{1}{c|}{71.05} &
  71.40~$\pm$~0.23 \\ \hline
QT-6 &
  90.63 &
  90.60 &
  \multicolumn{1}{c|}{99.20} &
  \multicolumn{1}{c|}{98.60} &
  98.80~$\pm$~0.22 &
  \multicolumn{1}{c|}{72.16} &
  \multicolumn{1}{c|}{69.02} &
  70.61~$\pm$~0.45 \\ \hline
QT-7 &
  95.31 &
  95.29 &
  \multicolumn{1}{c|}{99.40} &
  \multicolumn{1}{c|}{98.80} &
  99.08~$\pm$~0.24 &
  \multicolumn{1}{c|}{70.40} &
  \multicolumn{1}{c|}{66.82} &
   68.76~$\pm$~1.31 \\ \hline
QT-8 &
  97.66 &
  97.63 &
  \multicolumn{1}{c|}{99.00} &
  \multicolumn{1}{c|}{97.40} &
  98.56~$\pm$~0.62 &
  \multicolumn{1}{c|}{67.92} &
  \multicolumn{1}{c|}{65.30} &
  66.40~$\pm$~1.94 \\ \hline
QT-9 &
  98.83 &
  98.80 &
  \multicolumn{1}{c|}{99.20} &
  \multicolumn{1}{c|}{98.60} &
  98.72~$\pm$~0.27 &
  \multicolumn{1}{c|}{68.70} &
  \multicolumn{1}{c|}{63.55} &
  66.28~$\pm$~1.97 \\ \hline
QT-10 &
  99.41 &
  99.39 &
  \multicolumn{1}{c|}{99.00} &
  \multicolumn{1}{c|}{97.00} &
  98.08~$\pm$~0.70 &
  \multicolumn{1}{c|}{67.40} &
  \multicolumn{1}{c|}{61.82} &
  65.84~$\pm$~2.03 \\ \hline
\end{tabular}%
% }
\end{table*}
\begin{figure}[t]
\centering
\includegraphics[width=3.4in]{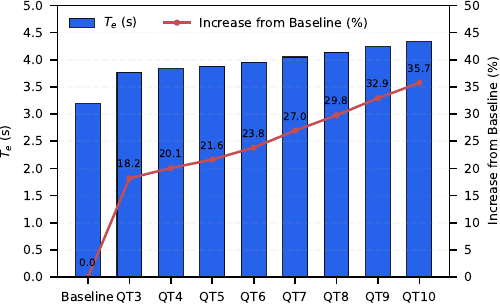}
    \caption{Training time per epoch of the baseline and QT variants.}
\label{time_comparison}
\end{figure}

Table~\ref{QT_full_comparison} presents the performance results of the QT framework on the UT-HAR and Widar3.0 datasets. For UT-HAR, we observe that all QT variants consistently improve or match baseline accuracy, even under aggressive parameter compression.
For example, QT‑5 and QT‑7 not only reduce the parameter count by approximately 81.22\% and 95.29\%, but also yield improved average accuracy (99.20 $\pm$ 0.22\% and 99.08 $\pm$ 0.24\%). These results suggest that the QT framework can successfully reduce parameter complexity without sacrificing model performance. In contrast, the performance on the more complex Widar3.0 dataset follows a different curve. QT-3 to QT-5 perform comparably or slightly better than the baseline, with QT-3 reaching 72.78 $\pm$ 0.22\% and QT-5 reaching 71.40 $\pm$ 0.23\%. Beyond QT-6, however, the performance starts to degrade, with QT-9 and QT-10 dropping to 66.28 $\pm$ 1.94\% and 65.84 $\pm$ 2.03\%, respectively. Furthermore, the variance in accuracy increases from 0.22 for QT-3 to 2.03 for QT-10, indicating a less stable model. This increasing variance, together with the drop in accuracy, hints that excessive compression may limit the performance of the QT framework. As the number of qubits increases, the dimensionality of the mapping network shrinks, thereby limiting its ability to learn good PQC probabilities and risking underfitting.

Regarding training efficiency, Fig.~\ref{time_comparison} shows that the baseline model on the UT-HAR dataset requires 3.2 seconds per epoch. In comparison, the QT framework exhibits a longer training time per epoch across all settings. Specifically, when increasing the number of qubits from 3 to 10, the training time increases gradually. For instance, QT-3 has an 18.2\% overhead compared to the baseline, and QT-10 has a 35.7\% increase. Although this introduces extra training cost, the gain in parameter efficiency makes it especially suitable for resource-constrained devices. Moreover, this additional training cost is incurred only during offline training and does not impact inference efficiency.

These outcomes provide a comprehensive  evaluation of the QT framework on classification performance, and training cost. The main advantage of QT is that it can vastly reduce the parameters without sacrificing performance, (and in some cases increases effectiveness) in classification.  A smaller qubit configuration (e.g., QT-3 to QT-5) exhibits moderate compression with low accuracy loss, while a larger number of qubit configurations (QT-8 to QT-10) achieves more compression but at the cost of reduced accuracy, particularly on more complex datasets. For the UT-HAR dataset, it is found that QT-7 has the best trade-off, achieving a high classification accuracy of 99.08 $\pm$ 0.24\% with a parameter reduction score of over 95\% and acceptable training overhead. In contrast, for the more complex Widar3.0 dataset, QT-6 has the best trade-off, maintaining an accuracy of 70.61 $\pm$ 0.45\% while reducing parameters by more than 90\%.

\subsubsection{Comparison with Lightweight ResNet-18}
\begin{table}[t]
\centering
\caption{Comparison of QT-7 against classically trained lightweight ResNet-18 and a static hypernetwork}
\label{Compare_QT7_LWResNet}
\resizebox{0.95\columnwidth}{!}{%
\begin{tabular}{|c|cc|cc|}
\hline
\multirow{2}{*}{\textbf{Model}} & \multicolumn{2}{c|}{\textbf{UT-HAR}}                     & \multicolumn{2}{c|}{\textbf{Widar3.0}}                    \\ \cline{2-5} 
                       & \multicolumn{1}{c|}{\textbf{Parameters}} & \textbf{Acc (\%)}      & \multicolumn{1}{c|}{\textbf{Parameters}} & \textbf{Acc (\%)}       \\ \hline
Lightweight ResNet-18  & \multicolumn{1}{c|}{529,295}    & 97.60 $\pm$ 0.42 & \multicolumn{1}{c|}{525,958}    & 62.60 $\pm$ 0.46 \\ \hline
Hypernetwork           & \multicolumn{1}{c|}{547,323}    & 98.60 $\pm$ 0.75 & \multicolumn{1}{c|}{547,244}    & 63.50 $\pm$ 7.39 \\ \hline
QT-7                   & \multicolumn{1}{c|}{547,424}    & 99.08 $\pm$ 0.24 & \multicolumn{1}{c|}{528,830}    & 68.76 $\pm$ 1.31 \\ \hline
\end{tabular}%
}
\end{table}
To assess whether Q-MET's gains stem from the training framework rather than from architectural choice, we compare QT-7 against a classically trained lightweight ResNet-18. Specifically, we reduce the number of channels in every convolutional layer of ResNet-18 by a width multiplier of 0.22, meaning each layer retains only 22\% of its original channel count. This yields a slim ResNet-18 with approximately 529k parameters, closely matched to the QT-7 parameter budget, ensuring a fair parameter-controlled comparison. As reported in Table~\ref{Compare_QT7_LWResNet}, the lightweight ResNet-18 achieves 97.60 $\pm$ 0.42\% on UT-HAR and 62.60 $\pm$ 0.46\% on Widar3.0, whereas QT-7 achieves 99.08 $\pm$ 0.24\% and 68.76 $\pm$ 1.31\%, respectively. The accuracy gap widens from 1.48\% on UT-HAR to 6.16\% on Widar3.0, demonstrating that Q-MET's advantage is attributable to the training framework and not merely to parameter budget reduction.

\subsubsection{Comparison with classical hypernetworks}
To assess whether Q-MET's gains stem from the proposed quantum-assisted framework rather than from classical parameter generation, we compare QT-7 against a static hypernetwork~\cite{ha2016hypernetworks,ha2017hypernetworks}. This baseline is chosen for two reasons: it is a well-established hypernetwork formulation, and it was originally designed to generate the convolutional kernels of CNNs, making it directly compatible with our ResNet-18 model. The hypernetwork is configured to generate all parameters of the ResNet-18 model, and its embedding size is tuned so that its trainable parameter count matches that of QT-7, ensuring a fair comparison under an equal parameter budget. As shown in Table~\ref{Compare_QT7_LWResNet}, the hypernetwork achieves an average classification accuracy of 98.60 $\pm$ 0.75\% on UT-HAR and 63.50 $\pm$ 7.39\% on Widar3.0, whereas QT-7 attains 99.08 $\pm$ 0.24\% and 68.76 $\pm$ 1.31\%, respectively. The accuracy gap widens from 0.48\% on UT-HAR to 5.26\% on Widar3.0. Moreover, the hypernetwork exhibits much higher run-to-run variance on Widar3.0 (a standard deviation of $\pm$7.39 compared with $\pm$1.31 for QT-7), indicating less stable training. These results indicate that Q-MET's advantage is attributable specifically to the proposed quantum-assisted framework rather than to the broader parameter-generation paradigm.

\begin{figure}[t]
\centering
\includegraphics[width=3.4in]{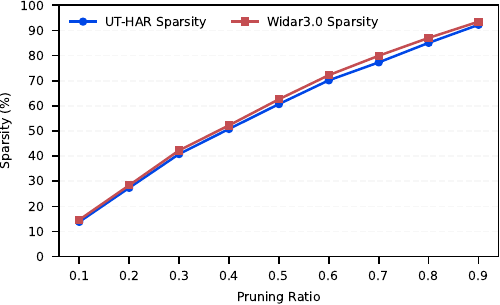}
    \caption{Sparsity of the model on the UT-HAR and Widar3.0 datasets across varying pruning ratios.}
\label{Sparsity_comparison}
\end{figure} 

\subsubsection{Impact of Pruning Ratios and Warm-Up Epochs on Q-MET}
Fig.~\ref{Sparsity_comparison} illustrates the sparsity of the ResNet-18 model on both datasets with different pruning ratios. Both models tend to have distinct and monotonic sparsity increases upon an increase in the pruning ratio. Importantly, the sparsity on both datasets exceeds 70\% when the pruning ratio is higher than 0.6, ranging up to 90\% at a pruning ratio of 0.9.

\begin{figure*}[!t]
  \centering
  \subfloat[]{    \includegraphics[height=1.7 in]{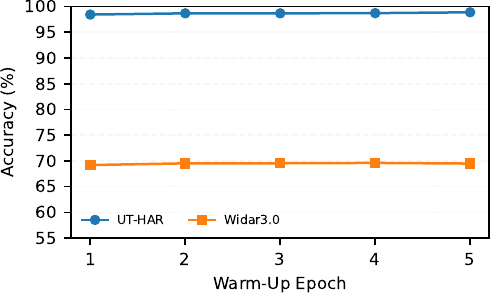}\label{fig:pruning_06}
  }
  \hspace{0.5cm}  % Adjust space between the images here
  \subfloat[]{    \includegraphics[height=1.7 in]{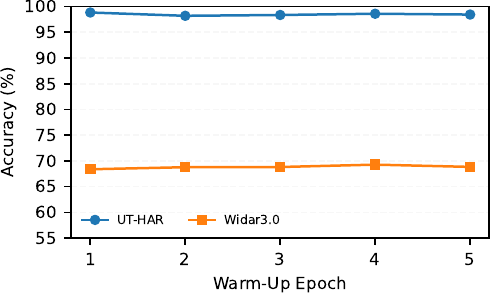}\label{fig:pruning_07}
  } \\
  \subfloat[]{    \includegraphics[height=1.7 in]{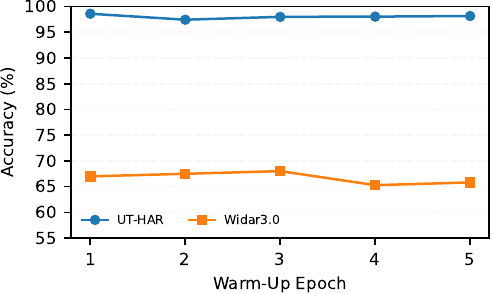}\label{fig:pruning_08}
  }
  \hspace{0.5cm}  % Adjust space between the images here
  \subfloat[]{    \includegraphics[height=1.7 in]{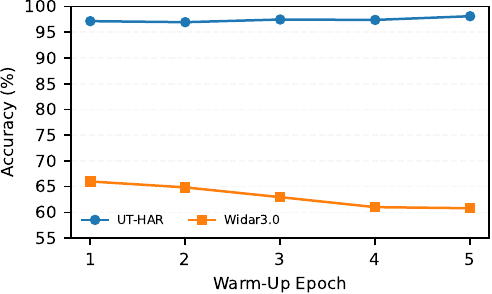}\label{fig:pruning_09}
  }
  \caption{Performance of Q-MET across warm-up epochs and pruning ratios on UT-HAR and Widar3.0 datasets. (a) Pruning ratio is 0.6. (b) Pruning ratio is 0.7. (c) Pruning ratio is 0.8. (d) Pruning ratio is 0.9.}
  \label{pruning_ratios_comparison}
\end{figure*}

Fig.~\ref{pruning_ratios_comparison} shows the impact of warm-up epochs on Q-MET performance under the selected pruning ratio. The duration of the Warm-Up is varied from 1 to 5 epochs to investigate whether using a longer warm-up period improves performance. This span is in line with common practice, where Warm-Up is generally applied for a few epochs to stabilize the early training process~\cite{Shen_2022_CVPR,peste2021ac}. The results suggest the number of warm-up epochs has little effect on classification accuracy in the two datasets. Accuracy on the UT-HAR dataset is stable and high in all warm-up settings with slight differences around 1\%. At a pruning ratio of 0.6, accuracy increases marginally from 98.44\% at one warm-up epoch to 98.88\% for five epochs, while at an aggressive pruning ratio (0.9), accuracy fluctuates around 96.96\% to 98.12\% without much indication of increases. For the Widar3.0 dataset, warm-up epochs generate small accuracy differences at moderate pruning ratios and no sustained performance improvement with the time spent warming up. In fact, at a pruning ratio of 0.9, accuracy drops from 66.01\% in one warm-up epoch to 60.81\% in five, suggesting that extended Warm-Up can be harmful with extreme sparsity. Due to the negligible gains and additional training overhead, a single warm-up epoch gives similar results to those for longer schedules and is thus elected as the optimal choice.

\subsubsection{Q-MET Performance}
\begin{figure*}[!t]
  \centering
  \subfloat[]{    \includegraphics[height=1.7 in]{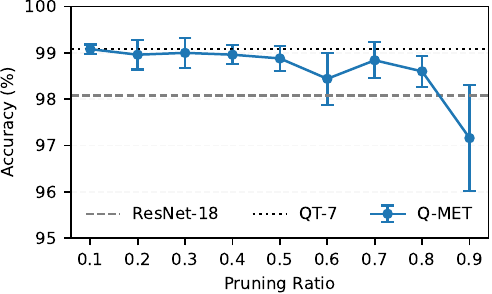}\label{fig:UT-HAR}
  }
  \hspace{0.5cm}  % Adjust space between the images here
  \subfloat[]{    \includegraphics[height=1.7 in]{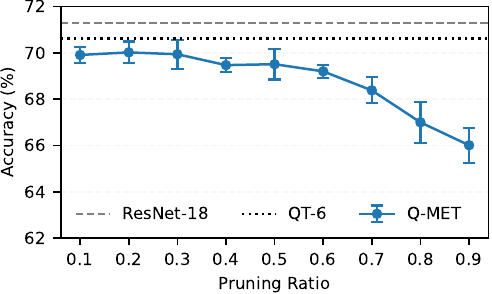}\label{fig:Widar}
  } 
  \caption{Performance of Q-MET across varying pruning ratios on UT-HAR and Widar3.0 datasets. (a) Q-MET on UT-HAR dataset. (b) Q-MET on Widar3.0 dataset.}
  \label{QMET_Acc_vs_ratio}
\end{figure*}

Following the previous analysis, we set the Warm-Up epoch to one and compare Q-MET against differing pruning ratios. Fig.~\ref{QMET_Acc_vs_ratio} displays the classification performance of Q-MET at different pruning ratios on the UT-HAR and Widar3.0 datasets with reference to ResNet-18 (Classical) and QT-based baselines. Overall, for both datasets, Q-MET provides competitive classification accuracy across many sparsity levels.

On the UT-HAR dataset, as shown in Fig.~\ref{QMET_Acc_vs_ratio}(a), Q-MET achieves high accuracy across pruning ratios from 0.1 to 0.8, with average accuracy remaining above 98.4\%. In particular, at moderate pruning levels (0.3 to 0.5), Q-MET reaches accuracies of 99.00\%, similar to or higher than the QT-7 baseline (99.08\%) and distinctly superior to the classical ResNet-18 baseline (98.08\%). Q-MET achieves over 98.44\% accuracy at a pruning ratio of 0.6 (where more than 70\% of parameters are removed), which signifies a small degradation in performance. A severe decline is evident at a pruning ratio of 0.9, where accuracy reaches 97.16\%, whilst variance levels rise to 1.15, further suggesting that extreme sparsity starts to limit model capacity. However, Q-MET can still score well against the unpruned ResNet-18 baseline, even under aggressive pruning.

For the Widar3.0 dataset, as shown in Fig.~\ref{QMET_Acc_vs_ratio}(b), Q-MET also demonstrates stable performance at moderate pruning ratios, delivering average accuracies around 70\% for pruning ratios up to 0.6. The results suggest that when using a 0.2 pruning ratio, Q-MET attains an accuracy peak of 70.02\%, remarkably close to the QT-6 baseline (70.61\%) and within a reasonable margin of the dense ResNet-18 baseline (71.29\%). Q-MET achieves about 69.5\% accuracy at pruning ratios of 0.3, 0.4, and 0.5. With the pruning ratio exceeding 0.6, there is a continuous gradual decrease in accuracy, which is 69.20\% at a pruning ratio of 0.6 and 66.01\% at 0.9. This trend reflects the higher sensitivity of Widar3.0 to model compression, consistent with the higher complexity of the Widar3.0 dataset.  The higher standard deviation at the higher pruning ratio is consistent with reduced stability under extreme sparsity.

\begin{figure}[!t]
  \centering
  \subfloat[]{\centering\includegraphics[width=3.2in]{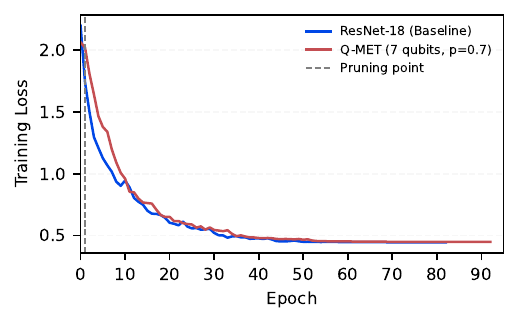}\label{train_loss}}\\[0.1cm]
  \subfloat[]{\centering\includegraphics[width=3.2in]{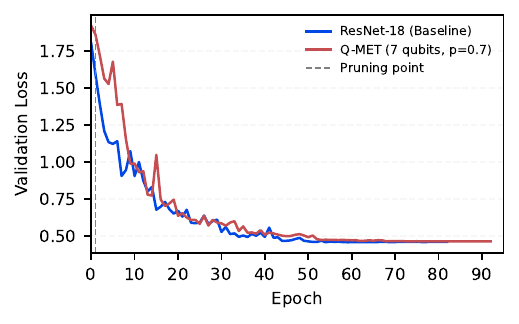}\label{val_loss}}
  \caption{ Convergence curves on the UT-HAR dataset, comparing the 
    baseline ResNet-18 and Q-MET (7 qubits, $p=0.7$). The dashed 
    line marks the pruning point at epoch 1. (a) Training loss. 
    (b) Validation loss.}
  \label{Train_val_loss}
\end{figure}

Fig.~\ref{Train_val_loss} shows the training loss and validation loss convergence curves on the UT-HAR dataset, comparing the baseline ResNet-18 trained with conventional backpropagation and Q-MET with 7 qubits and a pruning ratio of 0.7. Q-MET starts from a higher initial loss than the baseline, but after the early epochs both methods follow a comparable descent trajectory. The baseline converges and early-stops at epoch 83, while Q-MET converges and early-stops at epoch 93, despite using only 5\% of the original trainable parameters. A brief loss spike is observed in Q-MET immediately after the pruning point at epoch 1, where more than 70\% of the parameters are removed. However, the model recovers quickly within a few epochs and continues to decrease smoothly. Both methods reach similar final training loss values (approximately 0.45) and final validation loss values (approximately 0.46), confirming that Q-MET does not compromise convergence stability. 

\subsubsection{Comparison with Train-Then-Prune Pipeline}

To justify integrating pruning within the QT training process rather than applying it after training, we compare Q-MET against a train-then-prune (TTP) baseline in which QT-7 is first trained to convergence and then pruned with LAMP using the same criterion and pruning ratios. We evaluate three TTP variants to isolate the effect of fine-tuning after pruning: TTP$_0$ (no fine-tuning), TTP$_{10}$ (fine-tuning for 10 epochs), and TTP$_{50}$ (fine-tuning for 50 epochs). As shown in Table~\ref{QMET_and_TTP}, without any fine-tuning (TTP$_0$), the network fails completely at moderate-to-high pruning ratios. At a ratio of 0.7, accuracy drops to 9.60\,$\pm$\,1.96\% on UT-HAR and 3.71\,$\pm$\,2.77\% on Widar3.0. Adding more fine-tuning epochs recovers much of this loss. At the same ratio, TTP$_{50}$ reaches 98.80\% on UT-HAR and 64.74\% on Widar3.0, which is close to Q-MET's 98.84\% and 66.72\%. This recovery, however, requires 50 extra training epochs, while Q-MET needs no additional training after pruning. The benefit of Q-MET becomes clearer when more channels are removed. On UT-HAR at a ratio of 0.9, Q-MET still reaches 97.16\,$\pm$\,1.15\%, while TTP$_{50}$ falls to 92.64\,$\pm$\,4.60\% and becomes much less stable across runs. On the more difficult Widar3.0 dataset, Q-MET performs best at every pruning ratio, and the gap widens as pruning increases. At a ratio of 0.9, Q-MET reaches 65.73\% compared with 62.52\% for TTP$_{50}$ and only 6.10\% for TTP$_0$. Q-MET is also consistent, with standard deviations below 1.7 in all settings. These results support the design of Q-MET. By pruning during training, the QPG keeps adapting its parameter generation to the channels that remain active. The model therefore retains capacity that would be permanently lost if pruning were applied after training, and that fine-tuning can recover only in part.

\begin{table*}[t]
\centering
\caption{Comparison of Q-MET and TTP pipelines under varying pruning ratios.}
\label{QMET_and_TTP}
\resizebox{0.9\textwidth}{!}{%
\begin{tabular}{|c|c|c|c|c|c|c|c|c|}
\hline
\multirow{2}{*}{Pruning Ratio} & \multicolumn{4}{c|}{UT-HAR} & \multicolumn{4}{c|}{Widar3.0} \\ \cline{2-9} 
                               & Q-MET             & TTP$_0$           & TTP$_{10}$        & TTP$_{50}$        & Q-MET             & TTP$_0$           & TTP$_{10}$        & TTP$_{50}$        \\ \hline
QT-7 (Unpruned)                & \multicolumn{4}{c|}{99.08 $\pm$ 0.24} & \multicolumn{4}{c|}{68.76 $\pm$ 1.31} \\ \hline
0.1                            & 99.08 $\pm$ 0.10  & 84.32 $\pm$ 18.42 & 98.36 $\pm$ 0.59  & \textbf{99.16 $\pm$ 0.15}  & \textbf{68.43 $\pm$ 1.33}  & 62.08 $\pm$ 2.73  & 67.65 $\pm$ 0.89  & 66.17 $\pm$ 1.78  \\ \hline
0.2                            & 98.96 $\pm$ 0.32  & 46.88 $\pm$ 21.82 & 97.84 $\pm$ 0.86  & \textbf{99.16 $\pm$ 0.23}  & \textbf{69.00 $\pm$ 1.46}  & 47.27 $\pm$ 4.77  & 67.46 $\pm$ 1.10  & 66.16 $\pm$ 1.15  \\ \hline
0.3                            & 99.00 $\pm$ 0.33  & 24.88 $\pm$ 14.44 & 98.00 $\pm$ 0.55  & \textbf{99.28 $\pm$ 0.10}  & \textbf{68.96 $\pm$ 1.31}  & 19.19 $\pm$ 2.89  & 66.97 $\pm$ 1.22  & 65.89 $\pm$ 1.49  \\ \hline
0.4                            & 98.96 $\pm$ 0.20  & 10.84 $\pm$ 3.86  & 97.04 $\pm$ 1.22  & \textbf{99.36 $\pm$ 0.23}  & \textbf{68.46 $\pm$ 1.60}  & 7.46 $\pm$ 2.94   & 66.34 $\pm$ 0.92  & 65.50 $\pm$ 1.16  \\ \hline
0.5                            & 98.88 $\pm$ 0.27  & 9.72 $\pm$ 3.21   & 94.20 $\pm$ 2.86  & \textbf{99.12 $\pm$ 0.20}  & \textbf{68.71 $\pm$ 1.02}  & 5.14 $\pm$ 2.23   & 65.87 $\pm$ 0.51  & 65.04 $\pm$ 1.05  \\ \hline
0.6                            & 98.44 $\pm$ 0.56  & 9.60 $\pm$ 2.51   & 91.40 $\pm$ 2.75  & \textbf{99.16 $\pm$ 0.23}  & \textbf{67.15 $\pm$ 1.30}  & 1.47 $\pm$ 0.27   & 65.87 $\pm$ 0.91  & 64.94 $\pm$ 0.73  \\ \hline
0.7                            & \textbf{98.84 $\pm$ 0.39} & 9.60 $\pm$ 1.96   & 89.44 $\pm$ 1.34  & 98.80 $\pm$ 0.38  & \textbf{66.72 $\pm$ 1.19}  & 3.71 $\pm$ 2.77   & 64.22 $\pm$ 1.09  & 64.74 $\pm$ 0.78  \\ \hline
0.8                            & \textbf{98.60 $\pm$ 0.33} & 9.52 $\pm$ 1.98   & 80.84 $\pm$ 2.85  & 98.56 $\pm$ 0.41  & \textbf{66.70 $\pm$ 1.16}  & 3.59 $\pm$ 3.08   & 60.41 $\pm$ 2.81  & 63.84 $\pm$ 1.32  \\ \hline
0.9                            & \textbf{97.16 $\pm$ 1.15} & 9.36 $\pm$ 2.22   & 59.72 $\pm$ 5.39  & 92.64 $\pm$ 4.60  & \textbf{65.73 $\pm$ 0.66}  & 6.10 $\pm$ 3.07   & 52.78 $\pm$ 6.24  & 62.52 $\pm$ 1.79  \\ \hline
\end{tabular}%
}
\end{table*}

\subsection{Discussion}
The experimental results confirm that the Q-MET approach effectively balances training efficiency, inference performance, and model compactness, making it highly suitable for deployment in resource-constrained scenarios. During training, Q-MET leverages the QT framework to significantly reduce the trainable parameters, saving a large amount of memory while maintaining classification accuracy. In the UT-HAR dataset, Q-MET decreases the size of trainable parameters by about 95\%, which is accompanied by an improvement in classification accuracy from 98.08\% to 99.08\%. On the more challenging Widar3.0 dataset, Q-MET produces a 90\% decrease of trainable parameters and a minor decrease in accuracy from 71.29\% to 70.61\%. This small decrease in performance shows the trade-off between trainable parameters and model capacity on challenging datasets, even though it confirms that Q-MET can substantially reduce the trainable parameter count with minimal impact on recognition accuracy. Although QT framework introduces additional training time overhead compared to conventional training techniques, this is incurred only during offline training and is offset by substantial gains in memory and storage efficiency.

In terms of inference, Q-MET achieves substantial efficiency gains through structured pruning. Q-MET can achieve more than 85\% sparsity with a pruning ratio of 0.8 on the UT-HAR dataset and an accuracy of 98.60\%, within 1\% of the unpruned QT-7 model. Importantly, despite removing more than 85\% of the parameters, the Q-MET-trained pruned model still outperforms the unpruned model trained with the classical approach, which achieves an accuracy of 98.08\%. On the Widar3.0 dataset, it finds that models trained with the Q-MET framework still work effectively when about 75\% of the parameters have been trimmed (pruning ratio of 0.6), achieving results of 69.20\% in recognition accuracy, just 1.42\% lower than that achieved by an unpruned QT-6. Finally, this warm-up epoch analysis shows that one warm-up epoch is enough to obtain a steady performance under pruning, hence reducing the training costs overall and unnecessary computational overhead.
\section{Conclusion}
In this study, we proposed Q-MET, a novel memory-efficient end-to-end training framework for HAR. Q-MET leverages a hybrid quantum classical neural network architecture to efficiently train the classical ResNet-18 model. To optimize inference, we integrated LAMP directly into the QT process, effectively eliminating redundant channels and significantly reducing model size for deployment. Experimental results on the UT-HAR and Widar3.0 datasets demonstrate that Q-MET simultaneously enhances both training and inference efficiency.
On the UT-HAR dataset, Q-MET enabled training of the ResNet-18 model using only 5\% of the original trainable parameters (a 95\% reduction) and compressed the final model by approximately 85\%. Notably it achieved superior classification accuracy compared to the unpruned ResNet-18. On the more complex Widar3.0 dataset, Q-MET reduced the number of trainable parameters to 10\% of the original count and achieved 75\% model sparsity, with accuracy degradation limited to within 2\%. These findings demonstrate that Q-MET is a robust solution for enabling lightweight inference and memory-efficient training within resource-constrained HAR structures. Finally, our analysis of warm-up epochs revealed that a single warm-up epoch is sufficient to ensure stable pruning, thereby minimizing the computational overhead of the pre-training phase. Furthermore, since Q-MET is compatible with any compression technique, integrating it with other model-compression methods, such as quantization or hybrid pruning and quantization, is a promising direction for further reducing memory and storage costs. In addition, since HAR systems are often deployed in privacy-sensitive environments, incorporating security and privacy-preserving mechanisms into the Q-MET framework is another important direction. These include privacy-preserving training approaches such as federated learning, which avoid transferring raw CSI data to the cloud, as well as the evaluation and enhancement of the adversarial robustness of Q-MET-trained models against channel-level perturbations.
\bibliographystyle{IEEEtran}
\bibliography{IEEEabrv,Ref}
\end{document}